\documentclass[letterpaper, 10 pt, journal]{IEEEtran}

\IEEEoverridecommandlockouts

\usepackage{etextools}
\usepackage{amssymb}
\usepackage{float}
\usepackage{makeidx}
\usepackage{amsmath}
\usepackage{epsfig}
\usepackage{epsf}
\usepackage{psfrag}
\usepackage{verbatim}
\usepackage{color}
\usepackage{multirow}
\usepackage{tabularx}
\usepackage{booktabs}
\usepackage[tight,footnotesize]{subfigure}
\usepackage{array}
\usepackage{soul}
\usepackage{footnote}
\usepackage{cite}
\usepackage{dblfloatfix}

\usepackage{graphicx}
\graphicspath{{Figures}}
\DeclareGraphicsExtensions{.pdf,.png}
\usepackage{bigstrut}

\usepackage{algorithm}
\usepackage{algpseudocode}

\title{{Trajectory Deformations from Physical Human-Robot Interaction}}

\author{Dylan P. Losey, \textit{Student Member, IEEE}, and Marcia K. O'Malley, \textit{Senior Member, IEEE} 
\thanks{This work was funded in part by the NSF GRFP-1450681. \newline \indent
	The authors are with the Mechatronics and Haptic Interfaces Laboratory, Department of Mechanical Engineering, Rice University, Houston, TX 77005.
	{(e-mail: dlosey@rice.edu)}}}

\begin{document}
\maketitle

\begin{abstract}

Robots are finding new applications where physical interaction with a human is necessary: manufacturing, healthcare, and social tasks. Accordingly, the field of physical human-robot interaction (pHRI) has leveraged impedance control approaches, which support compliant interactions between human and robot. However, a limitation of {traditional} impedance control is that---despite provisions for the human to modify the robot's current trajectory---the human cannot affect the robot's future {desired} trajectory {through} pHRI. In this paper, we present an algorithm for physically interactive trajectory deformations which, when combined with impedance control, allows the human to modulate both the actual and desired trajectories of the robot. {Unlike related works, our method explicitly deforms the future desired trajectory based on forces applied during pHRI, but does not require constant human guidance.} We present our approach and verify that this method is compatible with traditional impedance control. Next, we use constrained optimization to derive the deformation shape. Finally, we describe an algorithm for real time implementation, and perform simulations to test the arbitration parameters. Experimental results demonstrate {reduction in the human's effort and improvement in the movement quality} when compared to {pHRI with} impedance control alone.

\end{abstract}


\section{Introduction}

	Physical human-robot interaction (pHRI) has become more pervasive as robots transition from structured factory floors to unpredictable human environments.  Today we can find applications of pHRI not only within manufacturing, but also for rehabilitation, surgery, training, and comanipulation.  In many of these situations, the human and robot are working collaboratively \cite{jarrasse2012}; both agents share a common goal, mutually respond to each other's actions, and provide assistance when needed \cite{bratman1992}.  Of course, while the human and robot may agree upon the goal they are trying to reach or the task they are attempting to perform, they might disagree on the trajectory that should be followed.  Within this context, impedance control---as originally proposed by Hogan \cite{hogan1985}---has traditionally been leveraged to relate interaction forces with deviations from the robot's desired trajectory.  Impedance control helps to provide compliant, safe, and natural robotic behavior \cite{santis2008}, and is currently regarded as the most popular control paradigm for pHRI \cite{haddadin2016}.  Unfortunately, while impedance control enables the human to modify the robot's \textit{actual} trajectory, it does not allow the human to interact with the robot's \textit{desired} trajectory.  Practically, this can cause humans to expend more effort when working to change the behavior of the robot, leading to {higher effort, or ``inefficient,"} human-robot collaboration \cite{jarrasse2008, corteville2007, li2014}. \par

	As a result, extensions of impedance control have been developed where the robot proactively moves along the human's desired trajectory \cite{corteville2007, erden2010, li2014}.  Under these techniques, the human dictates the desired trajectory and leads the interactions, while the robot estimates the human's intent and acts as a transparent follower.  Because the human guides the robot, however, the robot cannot meaningfully intervene towards reaching the goal or completing the task, and hence collaboration is restricted.  Alternatively, shared control methods {for comanipulation} can be used, where the human and robot dynamically exchange leader and follower roles \cite{mortl2012, li2015, kucukyilmaz2013, medina2015}.  {Both the human and robot are able to contribute to the robot's motion, and the robot's level of autonomy is adjusted by the shared control allocation.}  Although the robot can now meaningfully contribute within this shared control, we observe that the human is again unable to {directly} alter the robot's desired trajectory {through forces applied during pHRI.}  Hence, it may be beneficial to develop an approach which combines the advantages of both changing the desired trajectory and sharing control.  Under such a scheme, haptic interactions could become a bidirectional information exchange; the human physically conveys task-relevant modifications to the robot, while the robot's force feedback informs the human about the current desired trajectory. \par

\begin{figure}[t]

	\begin{center}
		\includegraphics[width=0.95\columnwidth]{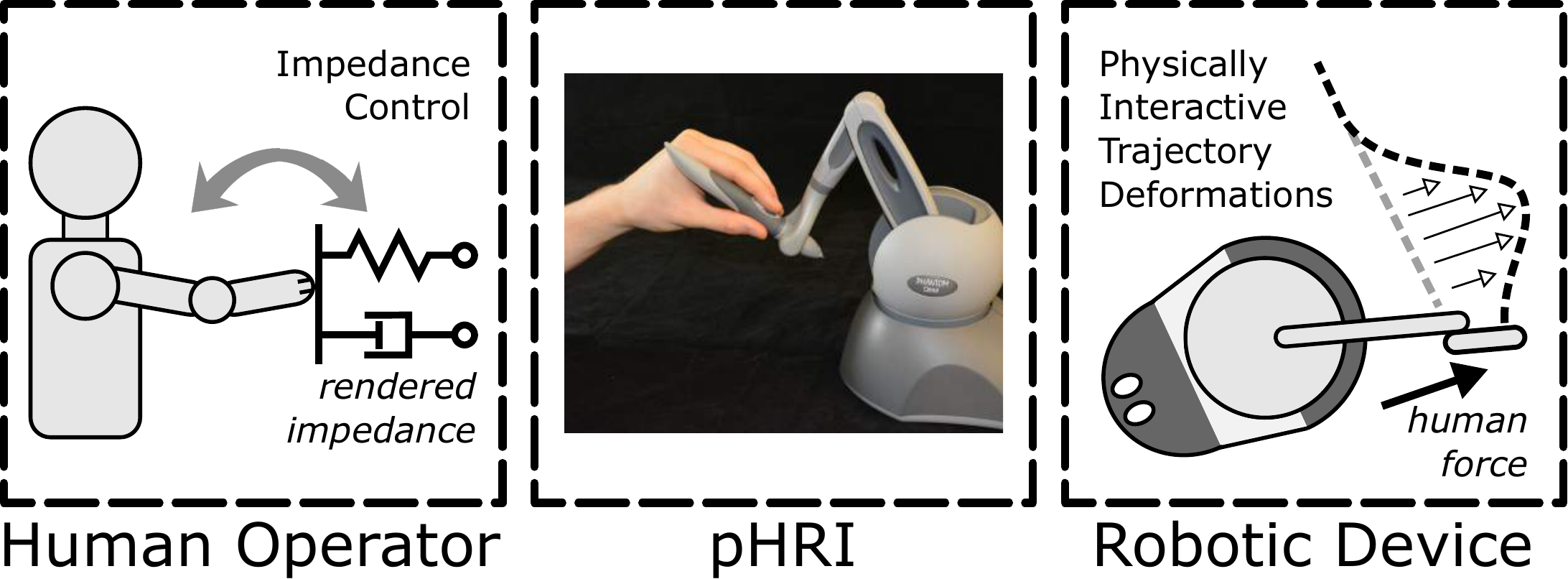}

		\caption{Schematic of the proposed framework for pHRI.  The human interacts with a {robot, which renders a desired impedance} (left).  In response to the human's applied force, the robot's future desired trajectory is updated using physically interactive trajectory deformations (right).}

		\label{fig:intro}
	\end{center}

	\vspace{-2em}

\end{figure}

	In this paper, we propose physically interactive trajectory deformations, which---when implemented alongside impedance control---enables the human to simultaneously interact with the robot's actual and desired trajectories.  By applying input forces, the human operator not only experiences the rendered compliance of the impedance controller, but also continually modifies or deforms a \textit{future} segment of the desired trajectory (see Fig.~\ref{fig:intro}).  Because the deformed desired trajectory returns to the original desired trajectory after some fixed time interval, the human does not need to constantly guide the robot, and so the robot can also contribute towards completing the shared task.  {Our approach is intended for applications where the human wants to change \textit{how} a robot behaves through physical interaction, and this robot is coupled to either a real or virtual environment.}  First, we derive constraints to ensure that the trajectory deformations are compatible with our impedance controller, and use constrained optimization to find a smooth, invariant deformation shape which resembles natural human motion.  Next, an algorithm implementing our approach in real time is presented, along with simulations demonstrating how the algorithm's parameters can be tuned to arbitrate between the human and robot.  Finally, we evaluate impedance control with physically interactive trajectory deformations by conducting human-subject experiments on a haptic device.  The results indicate that users exert less force and achieve better performance when interacting using the proposed method, as compared to a traditional impedance controller. \par

\section{Related Work} \label{sec:related}

	The problem we are considering and the solution we will pursue build on prior research from a variety of different fields.  In particular, for applications that involve pHRI, we will discuss recent work on tracking the human's desired trajectory and sharing control.  Outside of pHRI applications, we also overview trajectory changes during human-robot collaboration, and examine algorithms which can be used to deform the robot's desired trajectory in a smooth, human-like manner. \par

	As previously mentioned, the robot can actively follow the human's desired trajectory during pHRI; for example, \cite{corteville2007} uses a Kalman filter to track the human's desired timing of a point-to-point cooperative motion.  Erden and Tomiyama \cite{erden2010} measure the controller force as a means to detect the human's intent and update the robot's desired position---when the human stops interacting with the robot, the robot maintains its most recent position.  Similarly, Li and Ge \cite{li2014} employ neural networks to learn the mapping from measured inputs to the human's desired position, which the robot then tracks using an impedance controller.  {Although the human is able to change the robot's desired behavior, these methods require the human to guide the robot along their intended trajectory.} \par

	In contrast with \cite{corteville2007, erden2010, li2014}, shared control {for comanipulation} instead allows both the human and robot to dynamically exchange leader and follower roles \cite{kucukyilmaz2013, li2015, mortl2012, medina2015}.  In work by Li \textit{et al}. \cite{li2015}, game theory is used to adaptively determine the robot's role, such that the robot gradually becomes a leader when the human does not exert significant interaction forces.  Kucukyilmaz \textit{et al}. \cite{kucukyilmaz2013} have a similar criteria for role exchange, but find that performance decreases when visual and vibrotactile feedback informs the human about the robot's current role.  Medina \textit{et al}. \cite{medina2015} utilize stochastic data showing how humans have previously completed the task; at times where the robot's prediction does not match the human's behavior, prediction uncertainty and risk-sensitive optimal control decide how much assistance the robot should provide.  {We note that shared control methods such as \cite{jarrasse2012}, \cite{li2015}, and \cite{medina2015} leverage optimal control theory in order to modulate the controller feedback gains, but---unlike our proposed approach---they track a fixed desired trajectory.} \par

	{For shared control situations where the robot is continually in contact with an unpredictable environment---such as during a human-robot sawing task---Peternel et al. \cite{peternel2016} propose multi-modal communication interfaces, including force, myoelectric, and visual sensors.  By contrast, we consider tasks where the robot is attempting to avoid obstacles, and we focus on using pHRI forces without additional feedback.}  {Besides comanipulation, shared control has also been applied to teleoperation, where the human interacts with a haptic device, and that device commands the motions of an external robot.  In work by Masone et al. \cite{masone2012, masone2014}, the authors leverage haptic devices to tune the desired trajectory parameters of a quadrotor in real time.  These proposed adjustments are then autonomously corrected by the system to ensure path feasibility, regularity, and collision avoidance; afterwards, the haptic devices offer feedback about the resulting trajectory deformation.} \par

	Interestingly, even in settings where pHRI does not occur, other works have used the human's actions to cause changes in the robot's desired trajectory.  Mainprice and Berenson \cite{mainprice2013} present one such scheme, where the robot explicitly tries to avoid collisions with the human.  Based on a prediction of the human's workspace occupancy, the robot selects the desired trajectory which minimizes human-robot interference and task completion time.  Indeed, as pointed out by Chao and Thomaz \cite{chao2016}, if the human and robot are working together in close proximity---but wish to avoid physical contact---the workspace becomes a shared resource.  To support these methods, human-subject studies have experimentally found that deforming the desired trajectory in response to human actions objectively and subjectively improves human-robot collaboration \cite{lasota2015}.  However, it is not necessarily clear which trajectory deformation is optimal; as a result, there is interest in understanding how humans modify their own trajectories during similar situations.  Pham and Nakamura \cite{pham2015} develop a trajectory deformation algorithm which preserves the original trajectory's affine invariant features, with applications in transferring recorded human motions to humanoid robots. \par

	Finally, from a motion planning perspective, optimization methods can be used to find human-like and collision-free desired trajectories by iteratively deforming the initial desired trajectory.  For example, in work on redundant manipulators by Brock and Khatib \cite{brock2002}, an initial desired trajectory from start to goal is given, and then potential fields are used to deform this trajectory in response to moving obstacles.  More recently, Zucker \textit{et al}. developed CHOMP \cite{zucker2013}, an optimization approach which uses covariant gradient descent to find the minimum cost desired trajectory; each step down the gradient deforms the previous desired trajectory.  STOMP, from \cite{kalakrishnan2011}, generates a set of noisy deformations around the current desired trajectory, and then combines the beneficial aspects of those deformations to update the desired trajectory.  TrajOpt, from \cite{schulman2014}, uses sequentially convex optimization to deform the initial desired trajectory---the resulting deformation satisfies both equality and inequality constraints.  We observe that the discussed trajectory optimization schemes, \cite{brock2002, zucker2013, kalakrishnan2011, park2012, schulman2014}, are not intended for pHRI, but have been successfully utilized to share control during teleoperation \cite{dragan2013}. \par

\section{Nomenclature} \label{sec:nomenclature}

	We here introduce some of the variables that will be used in this paper.  For ease of notation, these variables are defined for a single degree-of-freedom (1-DoF) linear {robot}.  Our problem is first considered in an abstract manner---Section~\ref{sec:problem}---before we pursue a practical solution which can be implemented in real time---Section~\ref{sec:main}.  Accordingly, in Section~\ref{sec:problem}, we work with functions in order to be more accurate and complete.  Conversely, within Section~\ref{sec:main}, we approximate these functions using waypoint parameterizations.  The nomenclature discusses both cases as applicable. \par

\begin{itemize}[\usemathlabelsep]
	\item[$x_d^*$] $: \mathbb{R}^+ \rightarrow \mathbb{R}$ Original desired trajectory, assumed to be a smooth ($C^{\infty}$) function.
	\item[$x_d$] Desired trajectory, updated after each trajectory deformation.  In Section~\ref{sec:problem}, $x_d : \mathbb{R}^+ \rightarrow \mathbb{R}$ is a function.  In Section~\ref{sec:main}, $x_d$ is a set of waypoints.
	\item[$h$,$k$] $\in \mathbb{N}$ Discrete time variables.
	\item[$T$] $\in \mathbb{R}^+$ Sample period for the computer interface.
	\item[$\delta$] $\in \mathbb{R}^+$ Sample period for the desired trajectory.  Equivalently, the time between waypoints along $x_d$.
	\item[$r$] $\in \mathbb{Z}^+$ Ratio between $\delta$ and $T$.
	\item[$\tau_i$] $\in \mathbb{R}^+$ Time at which the current trajectory deformation starts.  Also, in Section~\ref{sec:main}, the time associated with the most recent waypoint along $x_d$.
	\item[$\tau_f$] $\in \mathbb{R}^+$ Time at which the current trajectory deformation ends.  Also, in Section~\ref{sec:main}, the time associated with some future waypoint along $x_d$.
	\item[$\tau$] $\in \mathbb{R}^+$ Length of time of the trajectory deformation.  An integer multiple of $\delta$.
	\item[$\gamma_d$] Segment of the desired trajectory between $\tau_i$ and $\tau_f$.  In Section~\ref{sec:problem}, $\gamma_d: [\tau_i, \tau_f] \rightarrow \mathbb{R}$ is treated as a function.  In Section~\ref{sec:main}, $\gamma_d \in \mathbb{R}^N$ is a vector of waypoints.
	\item[$\tilde{\gamma}_d$] Deformation of $\gamma_d$ with the same dual representations.
	\item[$\Gamma_d$] $: \mathbb{R} \times [\tau_i, \tau_f] \rightarrow \mathbb{R}$ Smooth family of trajectories.
	\item[$\Gamma_d^s$] $: \mathbb{R} \times [\tau_i, \tau_f] \rightarrow \mathbb{R}$ Trajectory within the smooth family of trajectories, shorthand for $\Gamma_d(s,t)$ with $s =$ constant.  In Section~\ref{sec:problem}, $\Gamma_d^0(t) = \gamma_d(t)$ and $\Gamma_d^1(t) = \tilde{\gamma}_d(t)$.
	\item[$V$] Variation of $\gamma_d$ used to find $\tilde{\gamma}_d$.  As before, in Section~\ref{sec:problem}, $V : [\tau_i, \tau_f] \rightarrow \mathbb{R}$ is a vector field along $\gamma_d$, and in Section~\ref{sec:main}, $V \in \mathbb{R}^N$ is a vector.
	\item[$N$] $\in \mathbb{Z}^+$ Number of waypoints along $\gamma_d$ or $\tilde{\gamma}_d$.
	\item[$R$] $\in \mathbb{R}^{N \times N}$ Matrix determining an inner product on $\mathbb{R}^N$.
	\item[$H$] $\in \mathbb{R}^N$ Shape of the optimal variation.
	\item[$\mu$] $\in \mathbb{R}^+$ Admittance of the optimal variation.
\end{itemize}

\section{Problem Statement} \label{sec:problem}

	In order to develop a framework for physically interactive trajectory deformations, we will specifically consider 1-DoF linear {robots}.  Restricting ourselves to 1-DoF keeps the notation simple, and enables us to pose the problem in a straightforward, instructive manner.  Later, in Section~\ref{sec:DoF}, we will show that the algorithm we ultimately derive can be independently applied to each DoF; hence, we do not lose any generality by now focusing on a single DoF.  The {robot---which could be thought of as a haptic device---}can accordingly be modeled as
\begin{equation} \label{eq:S2E1}
	m\ddot{x}(t) + b\dot{x}(t) = f_a(t) {+} f_h(t)
\end{equation}
where $m$ is a point mass and $b$ is the viscous friction constant.  The position of this {robot} is denoted by $x$, and the device is subject to two external forces: $f_a$, the force applied by the actuator, and $f_h$, the force applied by the human.  We would like to control $f_a$ so that the {robot} follows a desired trajectory, $x_d$, while rendering a virtual impedance, where this impedance consists of a desired stiffness, $k_d$, and a desired damping, $b_d$. \par

	A computer interface connects the {robotic device} to the virtual impedance.  Recalling that computers necessarily introduce analog-to-digital conversion \cite{diolaiti2006}, the virtual force at each sample time is given by
\begin{multline} \label{eq:S2E2}
	f_v(hT) = k_d\Big(x_d(hT) - x(hT)\Big) + \\ b_d\Big(\dot{x}_d(hT) - \dot{x}(hT)\Big)  \quad \forall h \in \mathbb{N}
\end{multline}
Here $h$ is a discrete time variable, which increases by one after each sample, and $T$ is the sample period for the computer interface.  Next, using a zero-order hold (ZOH) to return from digital-to-analog, the actuator force becomes
\begin{equation} \label{eq:S2E3}
	f_a(t) = f_v(hT)  \quad \forall t \in [hT, (h+1)T{\big)}
\end{equation}
Viewed together, (\ref{eq:S2E1}), (\ref{eq:S2E2}), and (\ref{eq:S2E3}) describe a standard 1-DoF linear haptic device under impedance control, while also capturing the effects of discretization.  For a more in-depth analysis of this haptic system, particularly in regards to passivity, we recommend \cite{diolaiti2006, abbott2005, colgate1994}. \par

	Our objective is to enable the human to intuitively and consistently change the desired trajectory, $x_d$, by applying forces, $f_h$, to the {robotic} device.  We argue that the human's input should not only affect the robot's current state through the impedance controller, but also interact with the robot's future behavior via trajectory deformations.  In order to more formally explore this concept, let us discuss some notation.  The desired trajectory is initialized as $x_d^*$, a smooth, $C^{\infty}$ function provided by the operator, where, if pHRI never occurs, $x_d(t) = x_d^*(t)$.  Each time the human physically interacts with the {robot}, however, the desired trajectory is modified or ``deformed,'' and $x_d$ is updated to include this deformation.  {The overall timing of the desired trajectory $x_d$ is assumed to be correct, and, while deformations can locally alter the speed of the robot, the total time required to complete the task will not be changed by these deformations.}  Although we will first focus on a single trajectory deformation, it should be understood that $x_d$ is continually modified in response to $f_h$ through an iterative process; we refer to this approach as physically interactive trajectory deformations. \par

\begin{figure}[t]

	\begin{center}
		\includegraphics[width=0.9\columnwidth]{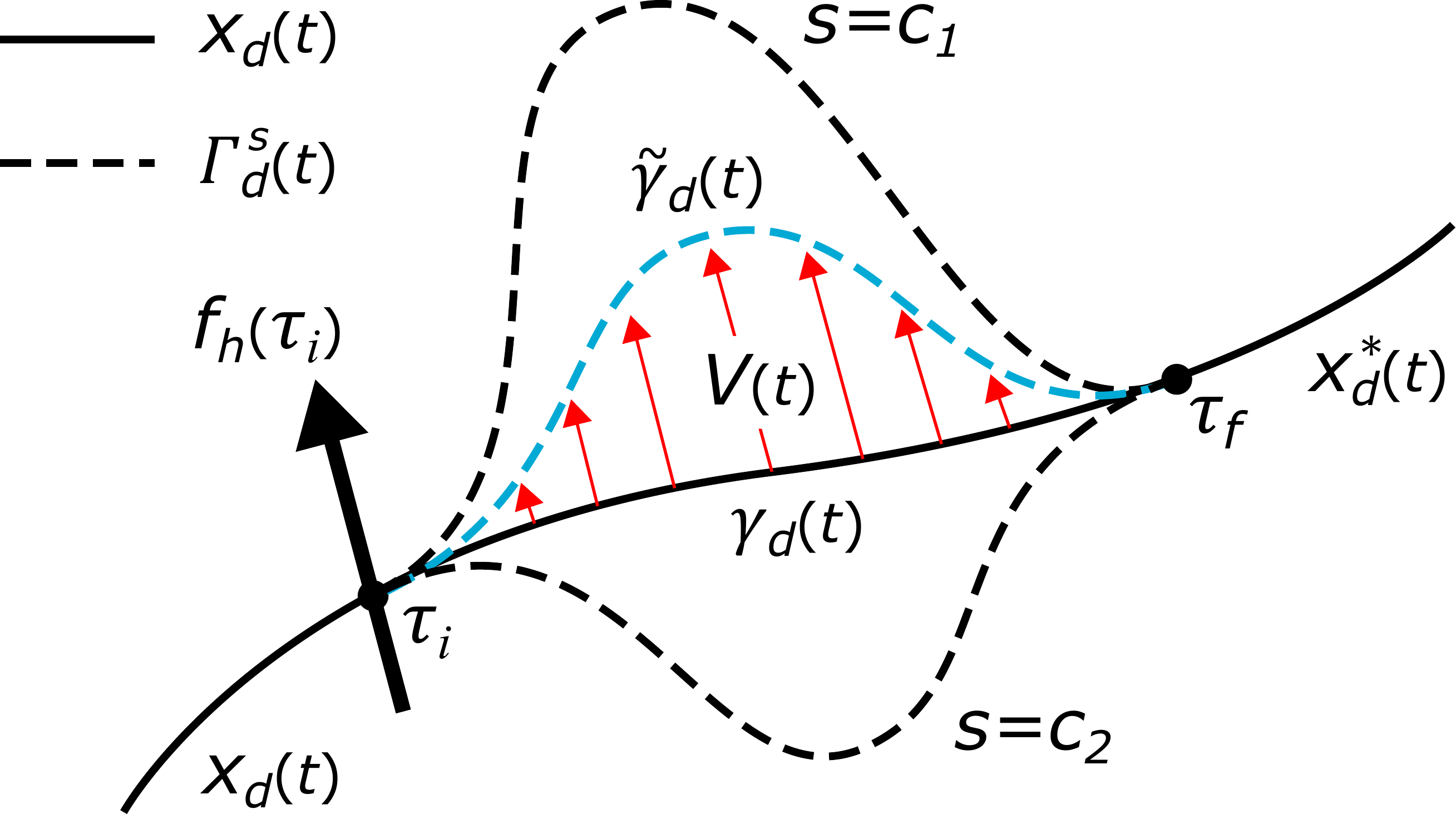}

		\caption{Diagram of the current trajectory deformation.  The segment of $x_d$ between times $\tau_i$ and $\tau_f$, labeled $\gamma_d$, is modified based on the human's force $f_h(\tau_i)$.  The resultant deformation is $\tilde{\gamma}_d$ (in blue).  The variation field $V$ (in red) can be used to obtain $\tilde{\gamma}_d$ from $\gamma_d$.  Other possible trajectory deformations, which are members of a smooth family of trajectories $\Gamma_d$, are also plotted (dashed lines).  We use $c_1$ and $c_2$ to denote constants.  Note that $\gamma_d$ is $\Gamma_d^0$, i.e., $s=0$, and $\tilde{\gamma}_d$ is $\Gamma_d^1$, i.e., $s=1$.  After $\tilde{\gamma}_d$ is determined, $x_d$ would next be updated such that $x_d(t) = \tilde{\gamma}_d(t)$ over the time interval $t \in [\tau_i, \tau_f]$.}

		\label{fig:variation}
	\end{center}

	\vspace{-2em}

\end{figure}

	For the current trajectory deformation, define $\tau_i$ to be the time at which $x_d$ starts to change, define $\tau_f$ to be a future time at which the current trajectory deformation ends, and define $\tau = \tau_f - \tau_i$ to be the duration of trajectory deformation.  Thus, if the human applies a force $f_h$ at an arbitrary time $\tau_i$, that human input can alter $x_d$ over the time interval $t \in [\tau_i, \tau_f]$.  We will define $\gamma_d : [\tau_i, \tau_f] \rightarrow \mathbb{R}$ as the restriction of $x_d$ to the interval $t \in [\tau_i, \tau_f]$, so that $\gamma_d$ refers only to the segment of the desired trajectory with which the human can currently interact.  Note that $\gamma_d(t) = x_d(t)$ when $t \in [\tau_i, \tau_f]$, but $\gamma_d$ is not defined outside this time interval.  Finally, let $\tilde{\gamma}_d$ denote the deformation of $\gamma_d$, which we will derive in Section~\ref{sec:main} by considering the human's force $f_h$ at sample time $\tau_i$.  Once $\tilde{\gamma}_d$ has been determined, $x_d$ is updated so that $x_d(t) = \tilde{\gamma}_d(t)$ over the interval $t \in [\tau_i, \tau_f]$.  After time $\tau_f$ the robot again follows its original desired trajectory, $x_d^*$, since the human has not yet interacted with this future portion of the trajectory---although he or she can in subsequent trajectory deformations.  To better visualize the setting, refer to Fig.~\ref{fig:variation}. \par

	Over the remainder of this section we identify the constraints on our choice of $\tilde{\gamma}_d$, and ensure that this trajectory deformation is compatible with our impedance controller.  We use the notation and definitions presented by Lee \cite{lee2006}.  Let $\Gamma_d(s,t)$ be a smooth family of trajectories---more commonly referred to as a smooth ``family of curves''---where $\Gamma_d : \mathbb{R} \times [\tau_i, \tau_f] \rightarrow \mathbb{R}$.  Intuitively, $\Gamma_d$ contains two collections of curves; changing $t$ when $s$ is constant allows us to move along a single trajectory, while changing $s$ when $t$ is constant allows us to move between multiple trajectories, evaluating each of these trajectories at time $t$.  For our purposes, the curves for which $s$ is constant are especially important, because each value of $s=$ constant will yield a different smooth trajectory over the time interval $t \in [\tau_i, \tau_f]$.  We write $\Gamma_d^s$ to denote these smooth trajectories with constant $s$, and define
\begin{equation} \label{eq:S2E4}
	 \Gamma_d^0(t) = \Gamma_d(0,t) = \gamma_d(t) \quad \forall t \in [\tau_i, \tau_f] 
\end{equation}
Hence, we have introduced $\Gamma_d^s$, where $\Gamma_d^0$ is $\gamma_d$, and all other values of $s$ give deformations of $\gamma_d$.  We observe that our use of $\Gamma_d^s$ is similar to the sets of homotopic paths employed for trajectory deformations in \cite{brock2002}.  Next, we utilize $\Gamma_d^s$ with the relationship from (\ref{eq:S2E2}) to obtain
\begin{multline} \label{eq:S2E5}
	F_v\big(\Gamma_d^s, hT\big) = k_d\Big(\Gamma_d^s(hT) - x(hT)\Big) + \\ b_d\Big(\partial_t\Gamma_d^s(hT) - \dot{x}(hT)\Big) \quad \forall h \in [\tau_i/T, \tau_f/T]
\end{multline}
Here $F_v$ provides the virtual force at sample time $hT$ when the desired trajectory is $\Gamma_d^s$, and $\partial_t$ denotes a partial derivative with respect to time.  As in (\ref{eq:S2E3}), the actuator force $F_a$ corresponding to $\Gamma_d^s$ is simply equal to the virtual force from (\ref{eq:S2E5}) evaluated at the most recent sample time
\begin{equation} \label{eq:S2E6}
	F_a\big(\Gamma_d^s, t\big) = F_v\big(\Gamma_d^s, hT\big) \quad \forall t \in [hT, (h+1)T{\big)}
\end{equation}
By construction, when $s=0$, we know that $F_a\big(\Gamma_d^s, t\big) = f_a(t)$ over the time interval $t \in [\tau_i, \tau_f]$.  As compared to (\ref{eq:S2E2}) and (\ref{eq:S2E3}), our equations (\ref{eq:S2E5}) and (\ref{eq:S2E6}) provide more generalized expressions for the virtual and actuator forces over the space of possible trajectory deformations. \par

	We now constrain the actuator force to be continuous when the trajectory deformation starts, $\tau_i$, and when the trajectory deformation ends, $\tau_f$.  This constraint guarantees that the human and robot will not experience any discontinuities in force when changing the desired trajectory, and accordingly prevents the trajectory modifications from interfering with our impedance controller.  Indeed, continuous interaction forces have been observed in human-human dyads \cite{reed2008}, and \cite{noohi2016} created an accurate model for human-robot interaction forces by assuming that these forces were continuous.  Leveraging the notation we have developed, we therefore assert that---regardless of our choice of $s$---the actuator force resulting from the deformed trajectory, $F_a$ in (\ref{eq:S2E6}), and the actuator force resulting from the original trajectory, $f_a$ in (\ref{eq:S2E3}), are equivalent at times $\tau_i$ and $\tau_f$  
\begin{equation} \label{eq:S2E7}
	F_a\big(\Gamma_d^s, \tau_i\big) = f_a(\tau_i) \quad \forall s
\end{equation}
\begin{equation} \label{eq:S2E8}
	F_a\big(\Gamma_d^s, \tau_f\big) = f_a(\tau_f) \quad \forall s
\end{equation}
Since (\ref{eq:S2E7}) and (\ref{eq:S2E8}) are satisfied when $s=0$, and $F_a = F_v$ at sample times $\tau_i$ and $\tau_f$, we could equivalently state
\begin{equation} \label{eq:S2E9}
	\partial_s F_v\big(\Gamma_d^0, \tau_i\big) = \partial_s F_v\big(\Gamma_d^0, \tau_f\big) = 0
\end{equation}
Again, $\partial_s$ denotes a partial derivative with respect to $s$.  For clarity, we point out that functions involving $\Gamma_d^s(t) = \Gamma_d(s,t)$ are really functions of $t$ and $s$; hence statements like $\partial_s\Gamma_d^0(t)$ imply that we first take the partial derivative of $\Gamma_d(s,t)$ with respect to $s$, and then evaluate this result for $s=0$. \par

	{We have arrived at constraints on $F_v$ which ensure that the human will experience continuous force feedback from the impedance controller when deforming the desired trajectory.  Perhaps more usefully, by applying (\ref{eq:S2E9}) to the right-side of (\ref{eq:S2E5}), we can re-express these constraints in terms of  $\Gamma_d^s$, and identify constraints on the shape of the trajectory deformation.}  Let $\Gamma_d^s$ be linear in $s$
\begin{equation} \label{eq:S2E10}
	\Gamma_d^s(t) = \gamma_d(t) + sV(t) \quad \forall t \in [\tau_i, \tau_f]
\end{equation}
Define $\Gamma_d^1(t) = \tilde{\gamma}_d(t)$, i.e., the curve $\Gamma_d(s,t)$ for which $s = 1$ is our trajectory deformation.  We now use (\ref{eq:S2E10}) to write
\begin{equation} \label{eq:S2E11}
	\tilde{\gamma}_d(t) = \gamma_d(t) + V(t) \quad \forall t \in [\tau_i, \tau_f]
\end{equation}
where $V$, the variation, is a vector field along $\gamma_d$ such that
\begin{equation} \label{eq:S2E12}
	V(t) = \partial_s \Gamma_d^0(t) \quad \forall t \in [\tau_i, \tau_f]
\end{equation}
Furthermore, from \cite{lee2006} and (\ref{eq:S2E12}), the following expressions are equivalent to the derivative of $V$ with respect to $t$
\begin{equation} \label{eq:S2E13}
	\partial_s \Big(\partial_t \Gamma_d^0(t)\Big) = \partial_t \Big(\partial_s \Gamma_d^0(t)\Big) = \dot{V}(t)
\end{equation}
Now, if we let $V$ and $\dot{V}$ equal zero at the start and end of the trajectory deformation,
\begin{equation} \label{eq:S2E14}
	V(\tau_i) = V(\tau_f) = 0
\end{equation}
\begin{equation} \label{eq:S2E15}
	\dot{V}(\tau_i) = \dot{V}(\tau_f) = 0
\end{equation}
we have that (\ref{eq:S2E9}) is satisfied, and hence the actuator force of the {robot}, $f_a$, is continuous when transitioning between original and deformed trajectories. \par

	More intuitively, recalling (\ref{eq:S2E11}), our derived constraints state that $\tilde{\gamma}_d(t) = \gamma_d(t)$ and $\dot{\tilde{\gamma}}_d(t) = \dot{\gamma}_d(t)$ both at the start, $\tau_i$, and the end, $\tau_f$, of the trajectory deformation.  These constraints aren't entirely unexpected.  On the one hand, having fixed endpoint positions---our constraint (\ref{eq:S2E14})---has been a typical requirement for motion planning with trajectory optimization \cite{kalakrishnan2011, park2012, zucker2013}.  On the other hand, having fixed endpoint velocities---our constraint (\ref{eq:S2E15})---can be a requirement for other trajectory deformation algorithms \cite{pham2015}, and even for modeling human-robot interaction forces \cite{noohi2016}.  As a result of our formulation, we have revealed that finding $\tilde{\gamma}_d$ boils down to determining $V$, the variation field.  Of particular interest, selecting a variation field fulfilling (\ref{eq:S2E14}) and (\ref{eq:S2E15}) ensures that the resultant trajectory deformation is compatible with our impedance controller.  Clearly, there are many choices of $V$ that satisfy these constraints; for demonstration, one such choice is shown in Fig.~\ref{fig:variation}.  Because we would like to find the ``best" variation field in accordance with (\ref{eq:S2E14}) and (\ref{eq:S2E15}), however, we turn our attention to optimization theory.

\section{Optimal Trajectory Deformations} \label{sec:main}

\begin{figure}[t]

	\begin{center}
		\includegraphics[width=0.95\columnwidth]{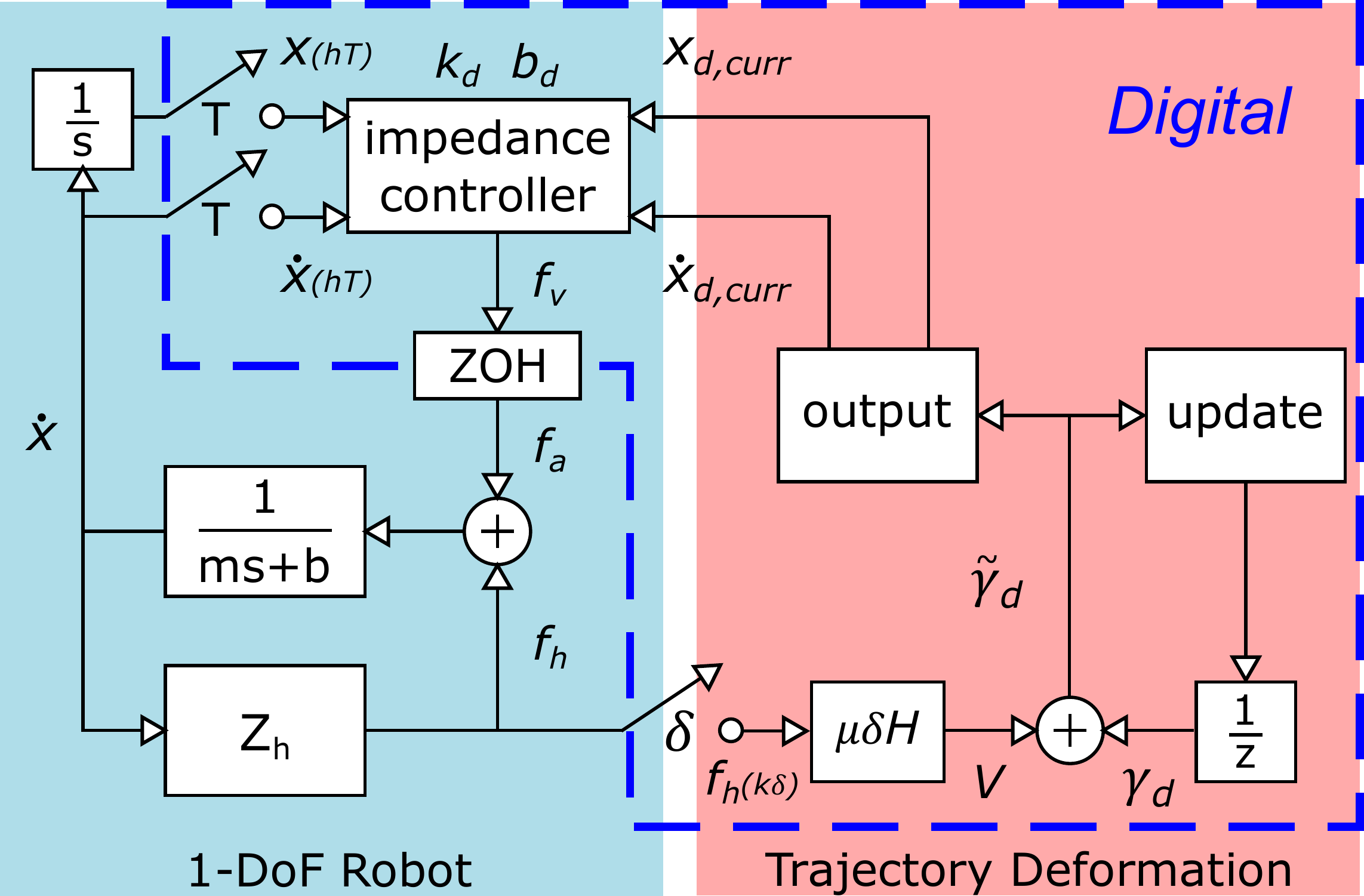}

		\caption{Block diagram of a 1-DoF {linear robot} (on left, blue) with physically interactive trajectory deformations (on right, red).  The digital portion (inside dashed line) is executed on a computer.  $Z_h$ is the human's impedance, ZOH stands for zero-order hold, and the device is modeled as a point mass ($m$) and viscous damper ($b$).  Recall that $1/z$ is a unit delay, and notice that sampling is occurring at two rates, with periods $T$ and $\delta$.}

		\label{fig:blocks}
	\end{center}

	\vspace{-2em}

\end{figure}

	Using the notation introduced in Section~\ref{sec:nomenclature} and the trajectory deformations described by Section~\ref{sec:problem}, we now present our method for physically interactive trajectory deformations.  At the conclusion of Section~\ref{sec:problem}, we argued that the optimal variation satisfying our endpoint constraints should be selected.  Hence, we will first describe the energy of the trajectory deformation, which is conveniently expressed using a waypoint trajectory parameterization.  Next, constrained optimization is leveraged to derive the variation $V$, and we explore how the shape of $V$ can be tuned to arbitrate between the human and robot.  We then present an algorithm that implements our results in real time, and combines impedance control with physically interactive trajectory deformations.  Finally, we extend the approach to $n$-DoF {robotic manipulators, and consider potential implementation issues.}  A block diagram of the scheme we are working towards is shown in Fig.~\ref{fig:blocks}.  The left panel of Fig.~\ref{fig:blocks} depicts equations (\ref{eq:S2E1}), (\ref{eq:S2E2}), and (\ref{eq:S2E3}) from Section~\ref{sec:problem}, while the steps and expressions in the right panel will be derived and explained below. \par

\subsection{Energy of the Trajectory Deformation} \label{sec:energy}

	In order to choose the optimal trajectory deformation, we first need to develop an energy function that maps trajectory deformations to their relative costs \cite{rao2009}.  Following the example of \cite{kalakrishnan2011, park2012, schulman2014, zucker2013}, we will formulate our energy function using a waypoint trajectory parameterization.  Let $x_d$ now be discretized into an arbitrary number of waypoints equally spaced in time, such that $\delta$ is the time interval between consecutive waypoints.  In particular, because the time duration of the segment of $x_d$ with which the human can interact is $\tau$, the number of waypoints along $\gamma_d$ and $\tilde{\gamma}_d$ is given by
\begin{equation} \label{eq:S3E1}
	N = \frac{\tau}{\delta} + 1
\end{equation}
We can therefore represent $\gamma_d$ and $\tilde{\gamma}_d$ as vectors of length $N$, where each element of these vectors is a waypoint.  For instance, referring back to the block diagram in Fig.~\ref{fig:blocks}, at $\tau_i = 0$ we initialize $\gamma_d$ to be
\begin{equation} \label{eq:S3E2}
	\gamma_d = [x_d^*(0),~x_d^*(\delta),~x_d^*(2\delta),~\ldots,~x_d^*(\tau)]^T
\end{equation}
Throughout the rest of Section~\ref{sec:main}, when we refer to $\gamma_d$, $\tilde{\gamma}_d$, or $V$, we will mean a vector in $\mathbb{R}^N$, and we will use the subscript $\gamma_{d,j}$ to refer to the $j$-th element of vector $\gamma_d$.  Applying this waypoint parameterization, the energy of the trajectory deformation is defined as
\begin{multline} \label{eq:S3E3}
	E(\tilde{\gamma}_d) = E(\gamma_d) + (\tilde{\gamma}_d - \gamma_d)^T ({-}\hat{F}_h) +\\  \frac{1}{2\alpha} (\tilde{\gamma}_d - \gamma_d)^T R (\tilde{\gamma}_d - \gamma_d)
\end{multline} 
Examining (\ref{eq:S3E3}), the trajectory deformation's energy is a summation of (a) the undeformed trajectory's energy, {(b) the work done by the trajectory deformation to the human}, and (c) the squared norm of $V$ with respect to the matrix $R$.  We will separately discuss (b) and (c). \par

	The desired trajectory should continually change in response to the human's applied force, $f_h$.  For this reason, we have included the second term in (\ref{eq:S3E3}), which will enable $f_h$ to affect the energy of the trajectory deformation.  Let $F_h \in \mathbb{R}^N$ be defined as the human's force applied along the variation vector $V$, where we recall from (\ref{eq:S2E11}) that $V = \tilde{\gamma}_d - \gamma_d$.  Thus, $F_{h,1}$ is the force applied at $V_1$, $F_{h,2}$ is the force applied at $V_2$, and so on.  While we know that $F_{h,1} = f_h(\tau_i)$, i.e., the current human force, we of course cannot know what $F_{h,2}$ through $F_{h,N}$ will be, since these forces occur at future times.  So as to develop a prediction of $F_h$---denoted as $\hat{F}_h$---we refer to work by Chipalkatty \textit{et al}. \cite{chipalkatty2013}. These authors considered a situation where, given the current human input, the robot attempts to both accomplish a high-level task and minimize the difference between its future actions and the human's future inputs.  After human experiments, the authors found that simpler prediction methods outperform more complex approaches, and are also preferred by users.  Like in \cite{chipalkatty2013} and \cite{noohi2016}, we therefore assume that future applied forces will be the same as the current input
\begin{equation} \label{eq:S3E4}
	\hat{F}_h = \vec{1} f_h(\tau_i), \quad \hat{F}_h \in \mathbb{R}^N
\end{equation}
Note that $\vec{1} \in \mathbb{R}^N$ is just a vector of all ones.  Substituting (\ref{eq:S3E4}) into (\ref{eq:S3E3}), {the second term is equivalent to the negative of the total waypoint displacement multiplied by the current force, i.e., the work done to the human}.  {Interestingly, if we compare (\ref{eq:S3E3}) to the energy function used for deriving the update rule within the CHOMP algorithm \cite{zucker2013}, we observe that $\nabla E(\gamma) = -\hat{F}_h$.  This implies that $\hat{F}_h$ is the direction of steepest descent, and, by deforming the desired trajectory in the direction of the human's applied force, we will move down the gradient of their energy function.  Thus, rather than attempting to explicitly learn $E$ in (\ref{eq:S3E3}), we are instead allowing the human to define the direction of steepest descent through pHRI.} \par

	The third term in (\ref{eq:S3E3}) ensures that the resulting trajectory deformation will seem natural to the human operator.  De Santis \textit{et al}. \cite{santis2008} explain that robotic trajectories for pHRI should resemble human movements.  In particular, the trajectory of human reaching movements can be accurately described by a minimum-jerk model \cite{flash1985}, and, more recently, the minimum-jerk model has been extended to account for human trajectory modifications \cite{meirovitch2016}.  Thus, we wish to assign lower energies to natural, minimum-jerk choices of $V = \tilde{\gamma}_d - \gamma_d$.  To achieve this end, let $A$ be a finite differencing matrix such that---when ignoring boundary conditions \cite{zucker2013}---we have $\dddot{V} = \delta^{-3} \cdot AV$
\begin{equation} \label{eq:S3E5}
	A = \begin{bmatrix} 1 & 0 & 0 & & 0 \\
				-3 & 1 & 0 &  & 0 \\
				3 & -3 & 1 & \cdots & 0 \\
				-1 & 3 & -3 & & 0 \\
				0 & -1 & 3 & & 0 \\
				0 & 0 & -1 & & 0 \\
				& \vdots & &\ddots & \vdots \\
				0 & 0 & 0 & & 1 \\
				0 & 0 & 0 & & -3 \\
				0 & 0 & 0 & \cdots & 3 \\
				0 & 0 & 0 & & -1 \\
	\end{bmatrix}, \quad A \in \mathbb{R}^{(N+3) \times N}
\end{equation}
Next, we define $R$, a positive definite and symmetric matrix formed using (\ref{eq:S3E5})
\begin{equation} \label{eq:S3E6}
	R = A^TA, \quad R \in \mathbb{R}^{N \times N}
\end{equation}
Both $A$ and $R$ are unitless quantities.  By construction,
\begin{equation} \label{eq:S3E7}
	\| \dddot{V} \|^2 = \dddot{V}^T\dddot{V} = \delta^{-6} \cdot V^TRV = \delta^{-6} \cdot \| V \|_R^2
\end{equation}
and so $R$ in (\ref{eq:S3E6}) determines an inner product on $\mathbb{R}^N$.  Unwinding these definitions, we additionally find that $\| V \|_R^2$ is proportional to the sum of squared jerks along the variation.  As a result, the third term in (\ref{eq:S3E3}) associates lower energies with natural shapes of $V$, i.e., those which minimize jerk. \par

\subsection{Constrained Optimization} \label{sec:optimization}

	Our definition of $E(\tilde{\gamma}_d)$ trades-off between the {work done to the human} and the variation's total jerk.  In order to find $V$, we would like to optimize $E(\tilde{\gamma}_d)$ subject to the constraints derived in Section~\ref{sec:problem}; however, (\ref{eq:S2E14}) and (\ref{eq:S2E15}) must first be re-written using our waypoint parameterization.  Let us introduce a matrix $B$, where
\begin{equation} \label{eq:S3E8}
	B = \begin{bmatrix} 1 & 0 & 0 & \cdots & 0 & 0 & 0 \\
				0 & 1 & 0 & \cdots & 0 & 0 & 0 \\
				0 & 0 & 0 & \cdots & 0 & 1 & 0 \\
				0 & 0 & 0 & \cdots & 0 & 0 & 1
		\end{bmatrix}, \quad B \in \mathbb{R}^{4 \times N}
\end{equation}
Using (\ref{eq:S3E8}), we concisely express (\ref{eq:S2E14}) and (\ref{eq:S2E15}) as
\begin{equation} \label{eq:S3E9}
	B(\tilde{\gamma}_d - \gamma_d) = 0
\end{equation}
Intuitively, the four constraints in (\ref{eq:S3E9}) guarantee that the first two and last two waypoints of $\tilde{\gamma}_d$ are the same as $\gamma_d$, which, in our prior terminology, implies that $V(t) =\dot{V}(t) = 0$ at the endpoints of the trajectory deformation, $\tau_i$ and $\tau_f$.  Combining (\ref{eq:S3E3}) and (\ref{eq:S3E9}), we now propose our optimization problem
\begin{equation} \label{eq:S3E10}
\begin{gathered}
	\text{minimize}~E(\tilde{\gamma}_d) \\
	\text{subject to} ~ B(\tilde{\gamma}_d - \gamma_d) = 0
\end{gathered}
\end{equation}
Solving (\ref{eq:S3E10}) both provides a ``best'' choice of $V$, and ensures that the resulting trajectory deformation does not interfere with the impedance controller.  Furthermore, we can augment this optimization problem by adding terms to the energy equation \cite{kalakrishnan2011, park2012, zucker2013}, and/or providing additional constraints \cite{schulman2014, zucker2013}.  Although we focus only on the simplest case, (\ref{eq:S3E10}), we do wish to point out that aspects of other optimal motion planners could also be incorporated within our approach. \par

	As demonstrated by \cite{dragan2015}, the optimization problem (\ref{eq:S3E10}) can be solved through a straightforward application of the method of Lagrange multipliers.  Let the Lagrangian be defined as
\begin{equation} \label{eq:S3E11}
	\mathcal{L}(\tilde{\gamma}_d, \lambda) = E(\tilde{\gamma}_d) + \lambda^T B(\tilde{\gamma}_d - \gamma_d)
\end{equation}
where $\lambda \in \mathbb{R}^4$ is a vector of Lagrange multipliers.  We use $\nabla_{X}$ to represent a gradient operator with respect to some vector $X$.  Following the procedure in \cite{rao2009}, the first-order necessity conditions for an extremum of $\mathcal{L}$ are given by
\begin{equation} \label{eq:S3E12}
	\nabla_{\tilde{\gamma}_d} \mathcal{L}(\tilde{\gamma}_d, \lambda) = {-}\hat{F}_h +  \frac{1}{\alpha}R (\tilde{\gamma}_d - \gamma_d) + B^T \lambda = 0
\end{equation}
\begin{equation} \label{eq:S3E13}
	\nabla_{\lambda} \mathcal{L}(\tilde{\gamma}_d, \lambda) = B(\tilde{\gamma}_d - \gamma_d) = 0
\end{equation}
After some algebraic manipulation of (\ref{eq:S3E12}) and (\ref{eq:S3E13}), and then substituting (\ref{eq:S3E4}) in for $\hat{F}_h$, we obtain
\begin{equation} \label{eq:S3E14}
	\tilde{\gamma}_d = \gamma_d {+} \alpha G f_h(\tau_i)
\end{equation}
We observe that the vector $G \in \mathbb{R}^{N}$ is formed using the identity matrix $I \in \mathbb{R}^{N \times N}$, $R$ from (\ref{eq:S3E6}), and $B$ from (\ref{eq:S3E8}).  Accordingly, $G$ is both unitless and constant, and can be completely defined once $N$ from (\ref{eq:S3E1}) is known
\begin{equation} \label{eq:S3E15}
	G = \Big( I - R^{-1}B^T(B R^{-1} B^T)^{-1} B \Big) R^{-1} \vec{1}
\end{equation}
Although we have found an extremum, it is not yet clear whether $\tilde{\gamma}_d$ in (\ref{eq:S3E14}) actually minimizes the energy function (\ref{eq:S3E3}).  To resolve this question, consider the Hessian of (\ref{eq:S3E3})
\begin{equation} \label{eq:S3E16}
	\nabla_{\tilde{\gamma}_d\tilde{\gamma}_d}E(\tilde{\gamma}_d) = \frac{1}{\alpha}R
\end{equation}
Because we have already established that $R$ is positive definite, and we define the constant $\alpha > 0$, we conclude that the Hessian (\ref{eq:S3E16}) is also positive definite.  This implies that $\tilde{\gamma}_d$ in (\ref{eq:S3E14}) minimizes $E(\tilde{\gamma}_d)$ subject to the equality constraints---solving the optimization problem (\ref{eq:S3E10}), as desired. \par

\subsection{Invariance and Arbitration}

\begin{figure}[b]

	\vspace{-1em}

	\begin{center}
		\includegraphics[width=0.9\columnwidth]{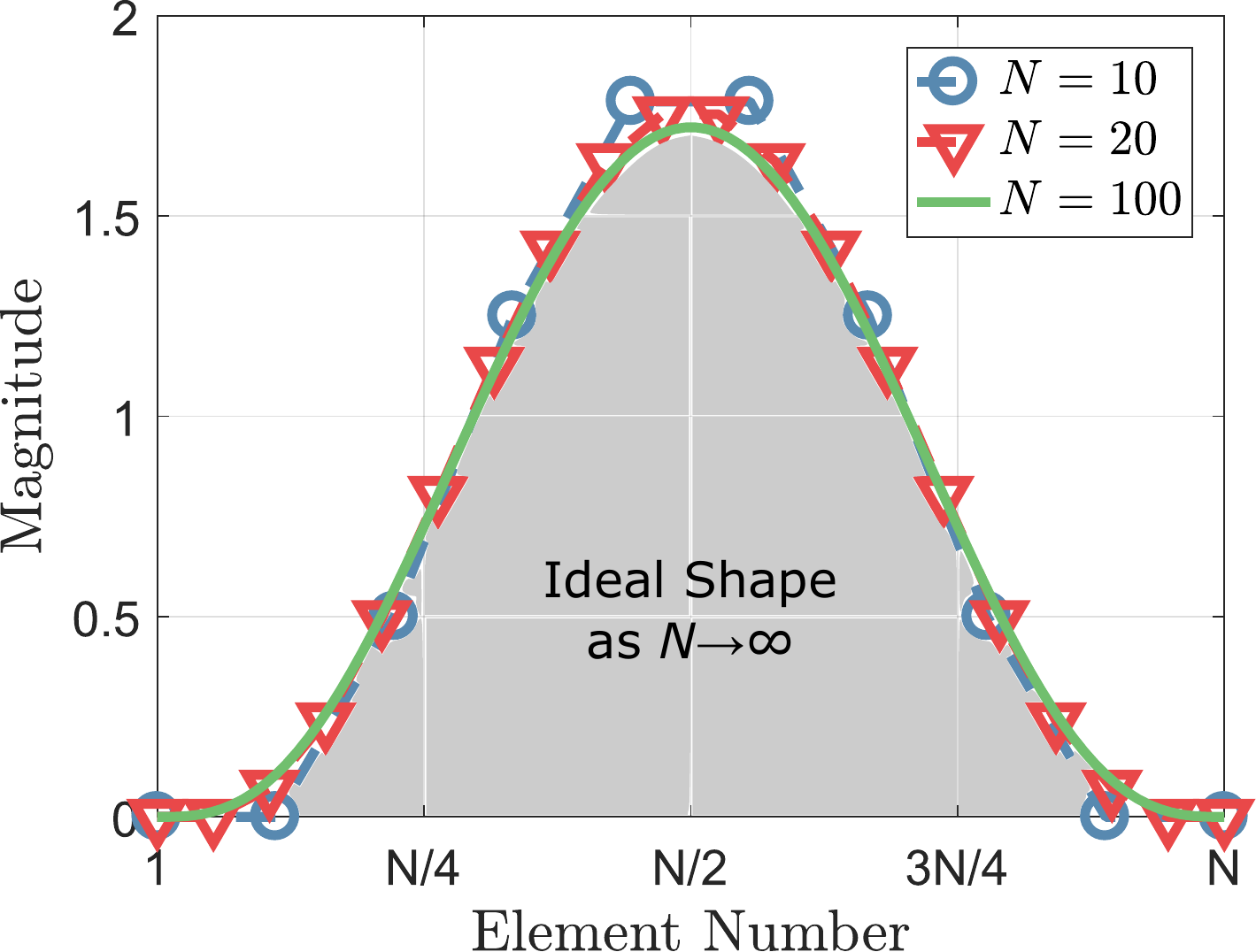}

		\caption{Shape of the optimal variation ($H$) vs. the number of waypoints ($N$).  The elements $1$ through $N$ of each vector $H \in \mathbb{R}^N$ are plotted, where we evaluated (\ref{eq:S3E17}) for $H$ and (\ref{eq:S3E15}) for $G$.  As $N$ increases, $H$ converges to the border of the shaded region.  Increasing the number of waypoints leads to a better approximation of the ideal shape, but does not alter the ideal shape itself.  Thus, the shape of $V$ is invariant with respect to $N$.  The first and last two elements of $H$ have zero magnitude due to our endpoint constraints (\ref{eq:S3E9}).}

		\label{fig:vector}
	\end{center}

	\vspace{-0.75em}

\end{figure}

	Comparing the result of our constrained optimization, (\ref{eq:S3E14}), to the original relationship between $\tilde{\gamma}_d$ and $\gamma_d$, (\ref{eq:S2E11}), we have revealed that $V = \alpha G f_h(\tau_i)$.  We next discuss our selection of the constant $\alpha$, where we have already constrained $\alpha > 0$.  Because the variation $V$ is a geometric object, the manner in which that object is represented should not alter its underlying shape \cite{zucker2013}; hence, the shape of $V$ should be invariant with respect to $N$, the number of waypoints.  Indeed, referring back to Fig.~\ref{fig:blocks}, our method for physically interactive trajectory deformations should additionally be invariant with respect to $1/\delta$, the rate at which trajectory deformations occur.  To achieve both types of invariance, we recall that $G$ in (\ref{eq:S3E15}) is a constant vector, and so we define
\begin{equation} \label{eq:S3E17}
	\alpha = \mu\delta \cdot \frac{\sqrt{N}}{\|G\|}, \quad H = \frac{\sqrt{N}}{\|G\|}G
\end{equation}
where $\mu$ is a positive scalar.  Then, substituting this choice of $\alpha$ into (\ref{eq:S3E14}), we reach our final equation for $\tilde{\gamma}_d$
\begin{equation} \label{eq:S3E18}
	\tilde{\gamma}_d = \gamma_d {+} \mu \delta H f_h(\tau_i)
\end{equation}
Due to the constraints found in Section~\ref{sec:problem}, the energy function from Section~\ref{sec:energy}, the optimization problem within Section~\ref{sec:optimization}, and the invariance just discussed, we have ultimately arrived at $V = \mu \delta H f_h(\tau_i)$.  Notice that the shape of $V$ is determined by the constant, unitless vector $H \in \mathbb{R}^N$.  Plots illustrating both the shape of our optimal variation and the invariance of $H$ with respect to $N$ can be seen in Fig.~\ref{fig:vector}. \par

	Our choice of $\mu$ in (\ref{eq:S3E18}) arbitrates how $V$ is impacted by $f_h$.  The user-specified parameter $\mu \in \mathbb{R}^+$ has SI units m/(N$\cdot$s), and is therefore analogous to an admittance.  Although there is some similarity between this parameter and the virtual springs in \cite{brock2002}, or the ``level of assistance'' in \cite{corteville2007}, our use of $\mu$ is most similar to $L$ in \cite{erden2010}.  Erden and Tomiyama leverage $L$, also a positive scalar, to relate the integral of the controller force to a change in the desired trajectory \cite{erden2010}.  Like this $L$, $\mu$ can be tuned to arbitrate between human and robot.  When $\mu$ is large, our method for physically interactive trajectory deformations arbitrates towards the human---smaller input forces cause larger deformations.  Conversely, as $\mu$ approaches zero, our method for physically interactive trajectory deformations arbitrates towards the robot---it becomes increasingly difficult for users to deform $x_d$. \par

\subsection{Algorithm for Implementation} \label{sec:algorithm}

	We have thus far focused on deriving an optimal variation for the current trajectory deformation.  Of course, after $\tilde{\gamma}_d$ is found using (\ref{eq:S3E18}), $x_d$ is updated to include $\tilde{\gamma}_d$, and the process iterates at the next trajectory deformation (see Fig.~\ref{fig:blocks}).  Here we present our algorithm to implement this iterative process in real time.  The algorithm has two loop rates; the computer interface---and impedance controller---have a sample period $T$.  Physically interactive trajectory deformations, on the other hand, has a sample period of $\delta$, since $\delta$ is the time interval between waypoints along $x_d$.  Let us define the ratio $r$ as $r = \delta/T$.  We assume that $r \in \mathbb{Z}^+$, since it is impractical to update the desired trajectory faster than the impedance controller can track $x_d$.  Our algorithm simplifies if $r = 1$, but, with $r = 1$, the algorithm could become too computationally expensive as $T$ decreases; this trade-off will be explored in Section~\ref{sec:simulations}.  Finally, we observe that $x_d$ is implicitly treated as a series of waypoints, where the desired position and velocity are held constant between those waypoints.  Explicitly, we will maintain only the most recent waypoint, $x_{d,curr}$, and the subsequent waypoint, $x_{d,next}$, which correspond to sample times $k\delta$ and $(k+1)\delta$.  From these two waypoints, discrete differentiation can be used to extract $\dot{x}_{d,curr}$.  Considering the ZOH between waypoints, $x_d$ as a function of time becomes
\begin{equation} \label{eq:S3E19}
	x_d(t) = x_{d,curr}(k\delta)  \quad \forall t \in [k\delta, (k+1)\delta{\big)}
\end{equation}
\begin{equation} \label{eq:S3E20}
	\dot{x}_d(t) = \dot{x}_{d,curr}(k\delta)  \quad \forall t \in [k\delta, (k+1)\delta{\big)}
\end{equation}
With these details in mind, we refer the reader to Algorithm~\ref{alg:myalg}.  This algorithm can be seen as complementary to Fig.~\ref{fig:blocks}, where the algorithm implements the ``digital'' portion of that block diagram.  To make Algorithm~\ref{alg:myalg} more concise, we have initialized $h$ and $k$ as $-1$, noting that, within the while loop, they are non-negative integers ($h,k \in \mathbb{N}$). \par

\begin{algorithm} [t!]
	\caption{method for combining impedance control with physically interactive trajectory deformations}
	\label{alg:myalg}
	\begin{algorithmic}

\State \textbf{Given:}
\State \quad \textbullet~Sampling periods: $T$, $\delta$
\State \quad \textbullet~User-selected parameters: $\tau$, $\mu$
\State \quad \textbullet~Original desired trajectory: $x_d^*(t)$
\State \quad \textbullet~Desired stiffness and damping: $k_d$, $b_d$
\State \textbf{Precompute:}
\State \quad \textbullet~Vector $H$ from (\ref{eq:S3E17}), where $N = \tau/\delta + 1$ from (\ref{eq:S3E1})
\State \textbf{Initialize:}
\State \quad \textbullet~$\gamma_{d} \gets [x_d^*(0),~x_d^*(\delta),~x_d^*(2\delta),~\ldots,~x_d^*(\tau)]^T$ \Comment{(\ref{eq:S3E2})}
\State \quad \textbullet~$h \gets -1$, $k \gets -1$
\State \quad \textbullet~$x_{d,curr} \gets \emptyset$, $\dot{x}_{d,curr} \gets \emptyset$
\newline
\While {$h~\text{is less than}~t_{stop}/T$} \Comment{$t_{stop}$ is the stop time}
	\State $h \gets h + 1$
	\If {$h~\text{is equal to}~r(k+1)$} \Comment{$r=\frac{\delta}{T}$}
		\State $k \gets k + 1$
		\State $\tau_i \gets k\delta$ \Comment{$\tau_f = \tau_i + \tau$}
		\State $f_h \gets $ \text{sampleForce}($\tau_i$)
		\State $\tilde{\gamma}_d \gets \gamma_d {+} \mu\delta H f_h$ \Comment{(\ref{eq:S3E18})}
		\State \textbf{output:}
		\State $x_{d,curr} \gets \tilde{\gamma}_{d,1}$ \Comment{$\tilde{\gamma}_{d,1} = x_{d,curr}$}
		\State $\dot{x}_{d,curr} \gets \frac{1}{\delta}(\tilde{\gamma}_{d,2} - \tilde{\gamma}_{d,1})$ \Comment{$\tilde{\gamma}_{d,2} = x_{d,next}$}
		\State \textbf{update:}
		\State $\gamma_{d} \gets [\tilde{\gamma}_{d,2},~\tilde{\gamma}_{d,3},~\ldots,~\tilde{\gamma}_{d,N}, ~x_d^*(\tau_f + \delta)]^T$
	\EndIf
	\State $x \gets $ \text{samplePosition}($hT$)
	\State $\dot{x} \gets $ \text{sampleVelocity}($hT$)
	\State $f_v \gets k_d(x_{d,curr} - x) + b_d(\dot{x}_{d,curr} - \dot{x})$ \Comment{(\ref{eq:S2E2})}
\EndWhile

	\end{algorithmic}
\end{algorithm}

\subsection{Extending to Multiple Degrees-of-Freedom} \label{sec:DoF}

	Now that physically interactive trajectory deformations has been established for a 1-DoF linear {robot}, we will show how this method can be extended to $n$-DoF.  In particular, we will consider a serial manipulator with forward kinematics $\boldsymbol{x} = \Phi(q)$, such that $q \in \mathbb{R}^n$ is the configuration in joint space, and $\boldsymbol{x} \in \mathbb{R}^m$ is the end-effector pose in task space ($m \leq 6$).  Here \textbf{bold} denotes the generalization of previously defined scalars into $m$-length vectors, so, for instance, $\boldsymbol{f_h} \in \mathbb{R}^m$ becomes the wrench applied by the human at the robot's end-effector.  From \cite{spong2006}, the joint space dynamics of the {robotic manipulator} are
\begin{equation} \label{eq:S3E21}
	M(q)\ddot{q} + C(q,\dot{q})\dot{q} + g(q) = u_r {+} J(q)^T \boldsymbol{f_h}
\end{equation}
where $M \in \mathbb{R}^{n\times n}$ is the inertia matrix, $C \in \mathbb{R}^{n\times n}$ contains the Coriolis and centrifugal terms, and $g \in \mathbb{R}^n$ is the gravity vector.  The robot actuators apply joint torques $u_r \in \mathbb{R}^n$, and $J(q) \in \mathbb{R}^{m \times n}$ is the robot's Jacobian matrix, which defines the mapping $\boldsymbol{\dot{x}} = J(q)\dot{q}$, and is assumed to be full rank. \par

	As before, we would like this {robotic manipulator} to track a desired end-effector trajectory, $\boldsymbol{x_d}$, while rendering a virtual stiffness and damping.  Hence, the virtual wrench---the wrench that the robot ideally applies in task space---can be written
\begin{multline} \label{eq:S3E22}
	\boldsymbol{f_v}(hT) = K_d\Big(\boldsymbol{x_d}(hT) - \boldsymbol{x}(hT)\Big) + \\ B_d\Big(\boldsymbol{\dot{x}_d}(hT) - \boldsymbol{\dot{x}}(hT)\Big) \quad \forall h \in \mathbb{N}
\end{multline}
Within (\ref{eq:S3E22}), $K_d \in \mathbb{R}^{m\times m}$ and $B_d \in \mathbb{R}^{m\times m}$ are the desired stiffness and damping matrices, respectively.  Let us define $K_d = \mbox{diag}(k_d^1, k_d^2,~\ldots, k_d^m)$ and $B_d = \mbox{diag}(b_d^1, b_d^2,~\ldots, b_d^m)$, where $k_d^j$ and $b_d^j$ are the desired stiffness and damping of the $j$-th task space coordinate.  Re-examining (\ref{eq:S3E22}) using these matrices, and introducing the notation $\boldsymbol{f_v} = [f_v^1, f_v^2,~\ldots, f_v^m]^T$, we see that the virtual force or torque $f_v^j$ is found by implementing Algorithm~\ref{alg:myalg} on the $j$-th coordinate in task space.  In other words, because there are no cross-terms in $K_d$ or $B_d$, $\boldsymbol{f_v}$ can be obtained by performing Algorithm~\ref{alg:myalg} separately with each of the $m$ task space coordinates, and then combining the results into an $m$-length vector.  We observe that $H$ from (\ref{eq:S3E17}), however, still only needs to be precomputed once. \par

	The control law in (\ref{eq:S3E21}) should be selected such that the human operator experiences $\boldsymbol{f_v}$, the virtual wrench, when interacting with our {robotic manipulator}.  {For simplicity of exposition}, we might consider the following controller
\begin{equation} \label{eq:S3E23}
	u_r = J(q)^T \boldsymbol{f_a} + u_{ff}
\end{equation}
Again applying a ZOH, $\boldsymbol{f_a}$ at time $t$ is equivalent to $\boldsymbol{f_v}$ evaluated at the most recent sample time, $hT$.  The feedforward torque vector, $u_{ff} \in \mathbb{R}^n$, could be used to either compensate for gravity or apply inverse dynamics \cite{spong2006}.  Of course, many other choices for $u_r$ are equally valid---e.g., see \cite{li2014, li2015,  medina2015}---but our principal finding here is not the design of (\ref{eq:S3E23}).  Instead, like the CHOMP \cite{zucker2013}, STOMP \cite{kalakrishnan2011}, and ITOMP \cite{park2012} algorithms, we have demonstrated that physically interactive trajectory deformations can straightforwardly scale to $n$-DoF manipulators.  More specifically, we must maintain a total of $m \cdot N$ waypoints, where $1 \leq m \leq 6$, and so our approach scales linearly with the dimensionality of the task space. \par

\subsection{{Unintended Interactions and Trajectory Constraints}} \label{sec:issues}

	{When implementing physically interactive trajectory deformations on single-DoF or multi-DoF robots, we must consider (a) unintended interactions and (b) trajectory constraints.  Up until this point, it has been assumed that $\boldsymbol{f_h}$ measures the external forces applied only by the human, and that all pHRI is intentional.  In practice, however, the robot could collide with some unknown obstacle, or the human might accidentally interact with the robot.  These unintended interactions alter $\boldsymbol{f_h}$, and, from Algorithm~\ref{alg:myalg}, change $\boldsymbol{x_d}$.  Therefore, to mitigate the effects of unintended interactions, we recommend filtering the external force $\boldsymbol{f_h}$.  For instance, \cite{erden2010} and \cite{kucukyilmaz2013} address similar issues by requiring the magnitude of $\boldsymbol{f_h}$ to exceed a predefined threshold for a given length of time, and then adjust the desired trajectory or shared control allocation based on that filtered $\boldsymbol{f_h}$.  During our human-subject experiments in Section~\ref{sec:experiments}, we will apply both a low-pass filter and minimum-force threshold before using $\boldsymbol{f_h}$ for physically interactive trajectory deformations.  As an aside, recall that these deformations only affect the future desired trajectory, and---since the stability of an impedance controller does not depend on the future desired trajectory \cite{spong2006}---we can conclude that our approach \textit{will not influence} impedance controller stability.} \par

	{After using Algorithm~\ref{alg:myalg} to deform $\boldsymbol{x_d}$, the updated desired trajectory may not satisfy some trajectory constraints; for instance, $\boldsymbol{x_d}$ could surpass the robot's joint limits, or collide with known obstacles.  Like we previously indicated, this problem can be addressed by incorporating additional costs or constraints within (\ref{eq:S3E10}), as shown by \cite{zucker2013, kalakrishnan2011, park2012, schulman2014}. A more straightforward solution, however, is simply to reject deformations which violate the trajectory constraints, and maintain the acceptable desired trajectory from the previous iteration.  After computing (\ref{eq:S3E18}) in Algorithm~\ref{alg:myalg}, we first check whether $\boldsymbol{\tilde{\gamma}_d}$ satisfies some margin of safety with respect to the robot's joint limits, free space, or other constraints.  If $\boldsymbol{\tilde{\gamma}_d}$ passes this check, proceed as usual; otherwise, reset $\boldsymbol{\tilde{\gamma}_d} = \boldsymbol{\gamma_d}$ before continuing.  Readers should be aware that---because of the computational time associated with collision checking---this additional procedure may decrease the rate at which physically interactive trajectory deformations are performed.} \par

\section{Simulations} \label{sec:simulations}

\begin{table}[t]

	\vspace{1.5em}

	\caption{Baseline Parameters Used in Simulations}
	\label{table:parameters}
	\begin{center}
		\begin{tabular}{l*{4}{c}}
			\textbf{Parameter} & {$T$ [s]} & {$\delta$ [s]} & {$\tau$ [s]} & {$\mu$ [m/N$\cdot$s]} \bigstrut \\ \hline
			\textbf{Value} & $10^{-3}$ & $10^{-2}$ & $1$ & $1$ \bigstrut[t] \\
		\end{tabular}
	\end{center}

	\vspace{-2em}

\end{table}

\begin{figure}[t]

	\begin{center}
		\includegraphics[width=0.825\columnwidth]{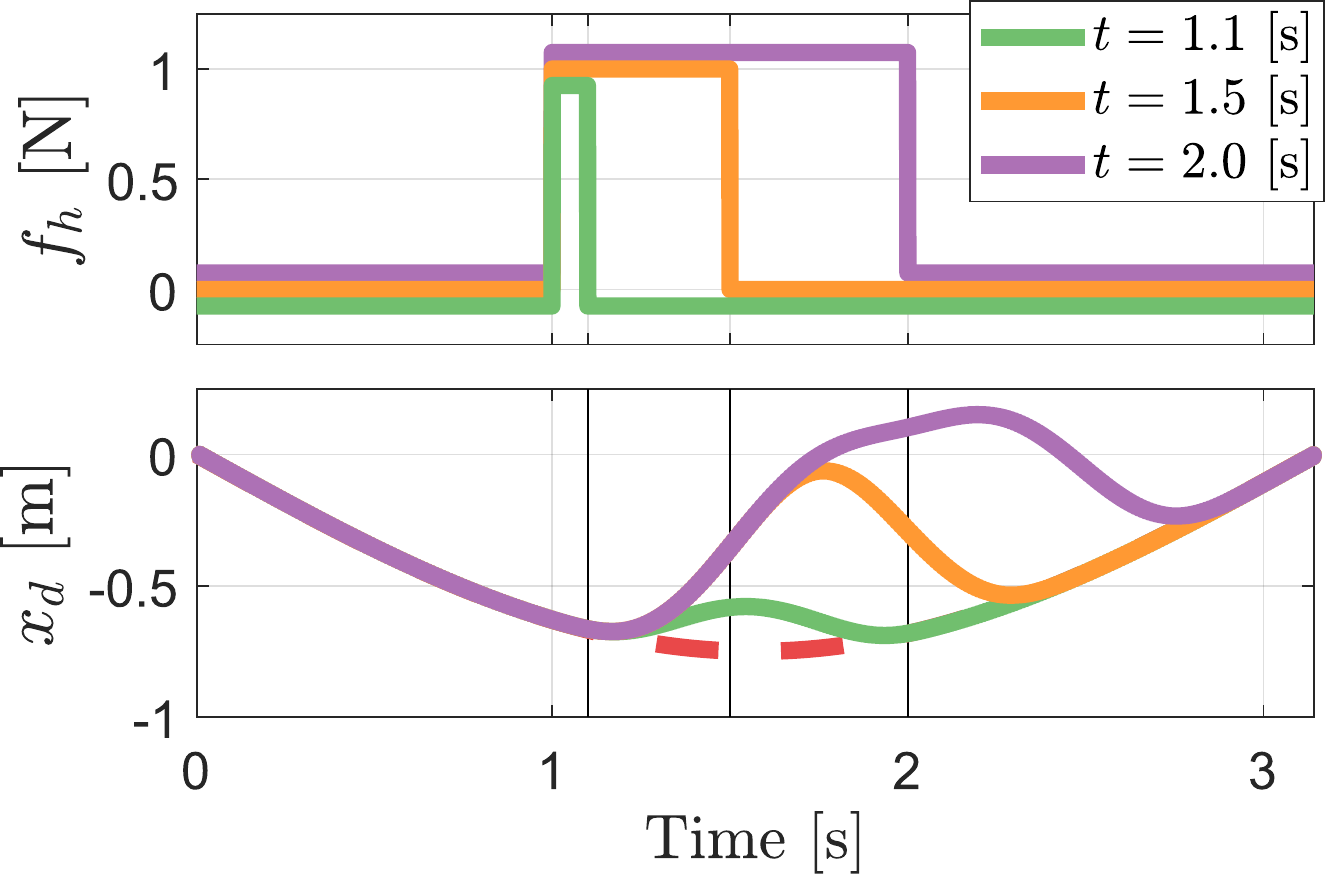}

		\caption{{Desired trajectory $x_d$ corresponding to different durations of applied force $f_h$. The magnitude of $f_h$ is the same in every plot, but we have added a visual offset for clarity. $x_d^*$ from (\ref{eq:S4E2}) is the dashed red line. The plot $t=1.1$~s shows $x_d$ after force is applied for $0.1$~s, the plot $t=1.5$~s shows $x_d$ after force is applied for $0.5$~s, and the plot $t=2.0$~s shows $x_d$ after force is applied for $1.0$~s. As the user applies force, $x_d$ iteratively deforms; these deformations aggregate smoothly over time.}}

		\label{fig:time}
	\end{center}

	\vspace{-0.75em}

\end{figure}

\begin{figure}[t]

	\begin{center}
		\includegraphics[width=0.825\columnwidth]{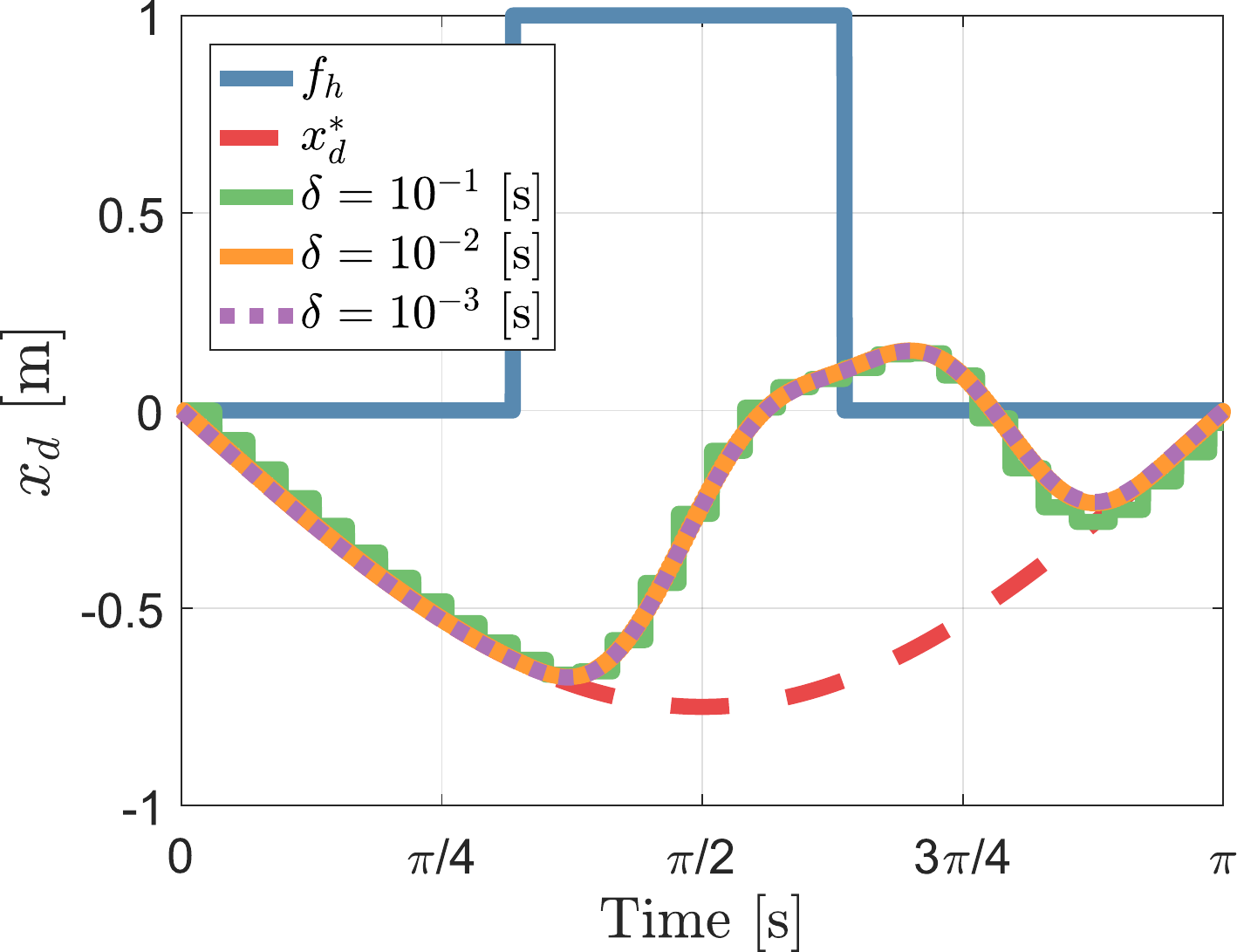}

		\caption{Effect of the sample period $\delta$ on the desired trajectory $x_d$.  Both $f_h$ from (\ref{eq:S4E1}), with units N, and $x_d^*$ from (\ref{eq:S4E2}), with units m, are plotted for reference.  Regardless of the value chosen for $\delta$, we follow the same underlying trajectory.  Hence, our method is invariant with respect to $\delta$.  Note that decreasing $\delta$ leads to a better approximation of the underlying trajectory, but it also increases $N$, the number of waypoints.  For example, when $\delta = 10^{-1}$ s we have $N = 11$, but when $\delta = 10^{-3}$ s, $N = 1001$.}

		\label{fig:delta}
	\end{center}

	\vspace{-2em}

\end{figure}

\begin{figure}[t]

	\begin{center}
		\includegraphics[width=0.825\columnwidth]{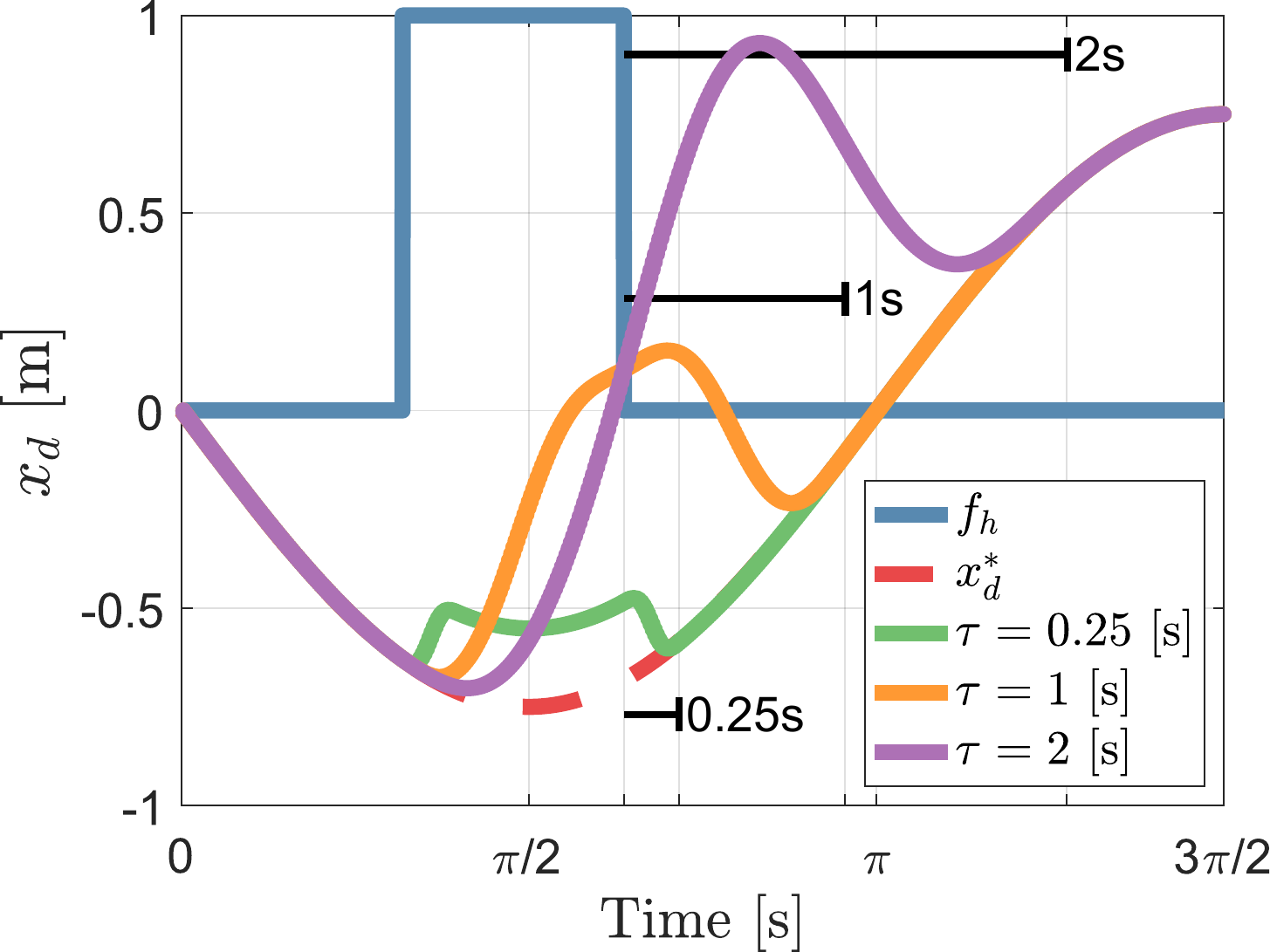}

		\caption{Effect of the parameter $\tau$ on the desired trajectory $x_d$.  Both $f_h$ from (\ref{eq:S4E1}), with units N, and $x_d^*$ from (\ref{eq:S4E2}), with units m, are plotted for reference.  We observe that $x_d$ smoothly returns to the original desired trajectory $\tau$ seconds after the human stops applying an input force (marked time intervals).  Increasing $\tau$ not only allows the human deform $x_d$ over a longer timescale, but it also increases the magnitude of the total deformation.}

		\label{fig:tau}
	\end{center}

	\vspace{-0.75em}

\end{figure}

\begin{figure}[t]

	\begin{center}
		\includegraphics[width=0.825\columnwidth]{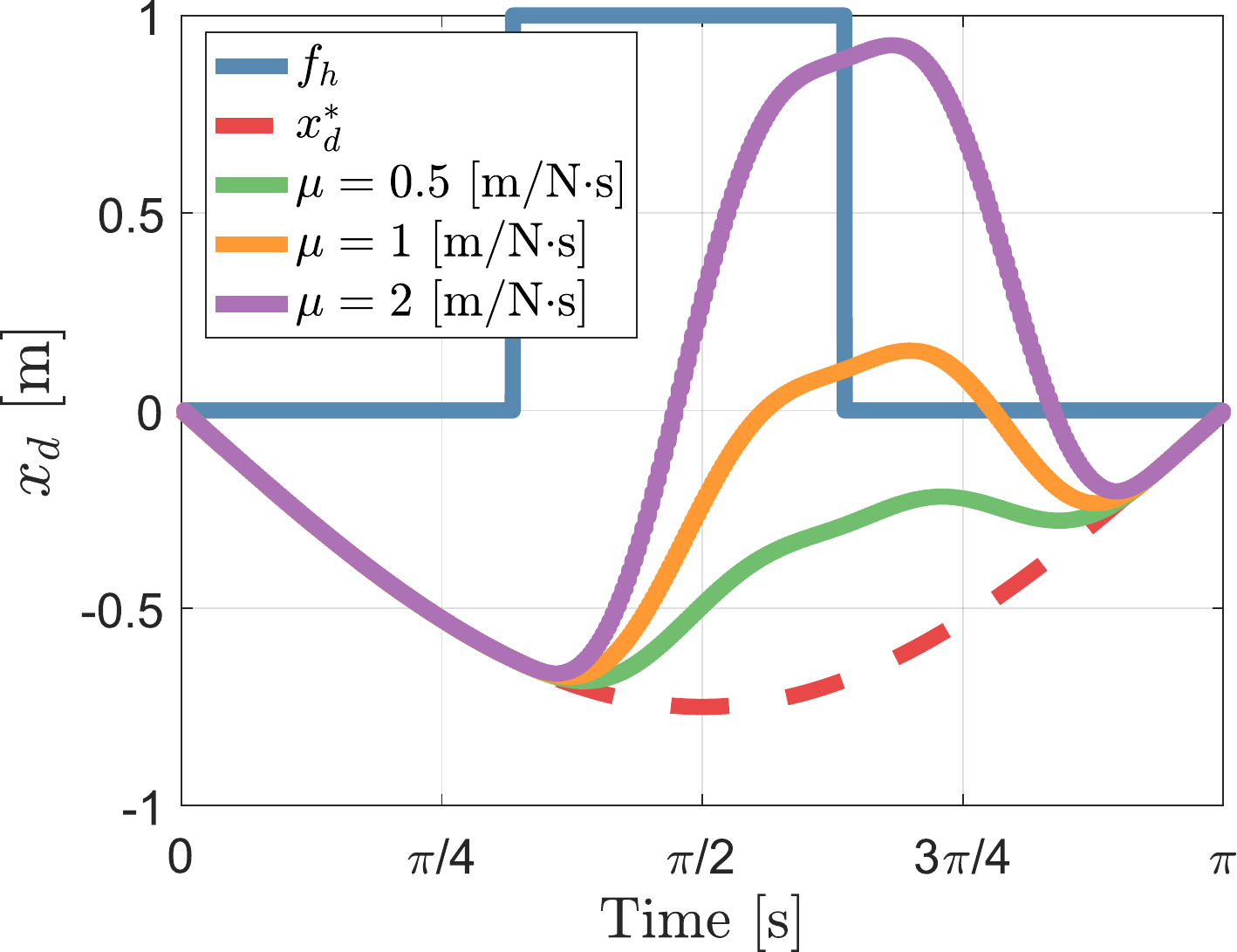}

		\caption{Effect of the parameter $\mu$ on the desired trajectory $x_d$.  Both $f_h$ from (\ref{eq:S4E1}), with units N, and $x_d^*$ from (\ref{eq:S4E2}), with units m, are plotted for reference.  Increasing $\mu$ arbitrates towards the human, so that the given force input causes a larger change in $x_d$.  The relationship between $\mu$ and the total deformation is not linear, although $\mu$ does linearly scale a single trajectory deformation.}

		\label{fig:mu}
	\end{center}

	\vspace{-2em}

\end{figure}

	To validate our method for physically interactive trajectory deformations and understand the effects of our user-specified parameters, we performed 1-DoF simulations in MATLAB (MathWorks).  These simulations demonstrate how the desired trajectory, $x_d$, deforms in response to forces applied by the human, $f_h$.  Unlike the hardware experiments in Section~\ref{sec:experiments}, here we will omit any {robotic device}.  Simulations were conducted using the ``Trajectory Deformation'' portion of the block diagram in Fig.~\ref{fig:blocks}, or, equivalently, Algorithm~\ref{alg:myalg}, excluding the final lines involving $x$, $\dot{x}$, or $f_v$.  In each simulation, we defined the human's force input to be a pulse function
\begin{equation} \label{eq:S4E1}
	f_h(t) = \begin{cases} 1\quad\mbox{if}~1 \leq t < 2 \\ 0\quad\mbox{otherwise} \end{cases}
\end{equation}
where $f_h$ is measured in Newtons.  Likewise, the original desired trajectory, $x_d^*$, was set as a sin wave
\begin{equation} \label{eq:S4E2}
	x_d^*(t) = {-}0.75 \cdot \sin(t)
\end{equation}
Both $x_d^*$ and $x_d$ were measured in meters.  So that readers can more easily reproduce our results, we have listed our simulation parameters in Table~\ref{table:parameters}.  Note that $k_d$ and $b_d$ are not necessary, since the virtual force $f_v$ is never computed.  Given these parameters, $f_h$ in (\ref{eq:S4E1}), and $x_d^*$ in (\ref{eq:S4E2}), we ran Algorithm~\ref{alg:myalg} with (\ref{eq:S3E19}) to find $x_d$ as a function of time.  Parameters $\delta$, $\tau$, and $\mu$ were then separately varied in different figures, where the parameters not currently being tested maintained their values from Table~\ref{table:parameters}.  Our results are shown in Figs.~\ref{fig:time}-\ref{fig:mu}.  {Fig.~\ref{fig:time} shows how $x_d$ smoothly and progressively deforms over time as the user continues to apply force $f_h$.}  Fig.~\ref{fig:delta} illustrates that changing $\delta$, the sampling period for physically interactive trajectory deformations, does not alter $x_d$.   Figs.~\ref{fig:tau} and \ref{fig:mu}, on the other hand, display the different ways in which user-specified parameters $\tau$ and $\mu$ can affect $x_d$.\par

	Viewing these figures together, it is clear that our method for physically interactive trajectory deformations enables the human to actively vary the desired trajectory.  When applying force $f_h=1$ N, the simulated user induces $x_d$ to iteratively deform, and, when no force is applied, $x_d$ again follows $x_d^*$ after $\tau$ seconds.  Importantly, we observe that the transitions between $x_d$ and $x_d^*$ are smooth; moreover, the deformed portions of the desired trajectory are also smooth.  {Note that $x_d$ deforms in the direction of $f_h$, maximizing the work done by the human's applied force.}  Besides validating Algorithm~\ref{alg:myalg}, these simulations have additionally resolved the previously discussed trade-off between $r=1$ and computational efficiency.  In order to have $r=1$, here we must choose $\delta = 10^{-3}$~s, which means we will have to constantly maintain $1000\tau+1$ waypoints.  As $\tau$ increases, this could eventually prevent real time implementation---however, even when $r=1$ and $\tau = 5$~s, i.e, $N =5001$, the simulations were still ten times faster than real time on a laptop computer.  Finally, Figs.~\ref{fig:tau} and \ref{fig:mu} provide some guidelines behind the selection of the user-defined parameters $\tau$ and $\mu$.  Increasing $\tau$ both causes the trajectory deformation to occur over a longer time interval, and increases the magnitude of the total deformation; therefore, to keep the total deformation magnitude constant, $\mu$ should be decreased as $\tau$ is increased. {In general, the choice of $\tau$ and $\mu$ may depend on the given robot, environment, and human-robot arbitration. During implementation, we recommend setting $\tau$ and $\mu$ at values close to zero, and then gradually increasing these values in accordance with our guidelines until the desired performance is achieved.} \par

\section{Experiments} \label{sec:experiments}

	To compare impedance control (IC) and impedance control with physically interactive trajectory deformations (IC-PITD), we conducted human-subject experiments on a multi-DoF haptic device, as shown in Fig.~\ref{fig:setup}.  Here the human and robot collaborated to follow a desired trajectory, where the human must intervene to help the robot avoid ``unknown'' obstacles.  {The physical robot was coupled to a virtual environment---which visualized these obstacles---and the virtual robot represented the position of the actual haptic device.}  We performed repetitions of the same tracking and obstacle avoidance task with IC and IC-PITD.  We hypothesized that (a) humans would {better} complete the task with IC-PITD, both in terms of {less} torque applied and {increased} movement smoothness, and (b) differences in tracking error and obstacle collisions between IC and IC-PITD would be negligible. \par

\subsection{Experimental Setup}

\begin{figure}[b]

	\vspace{0em}

	\begin{center}
		\includegraphics[width=0.6\columnwidth]{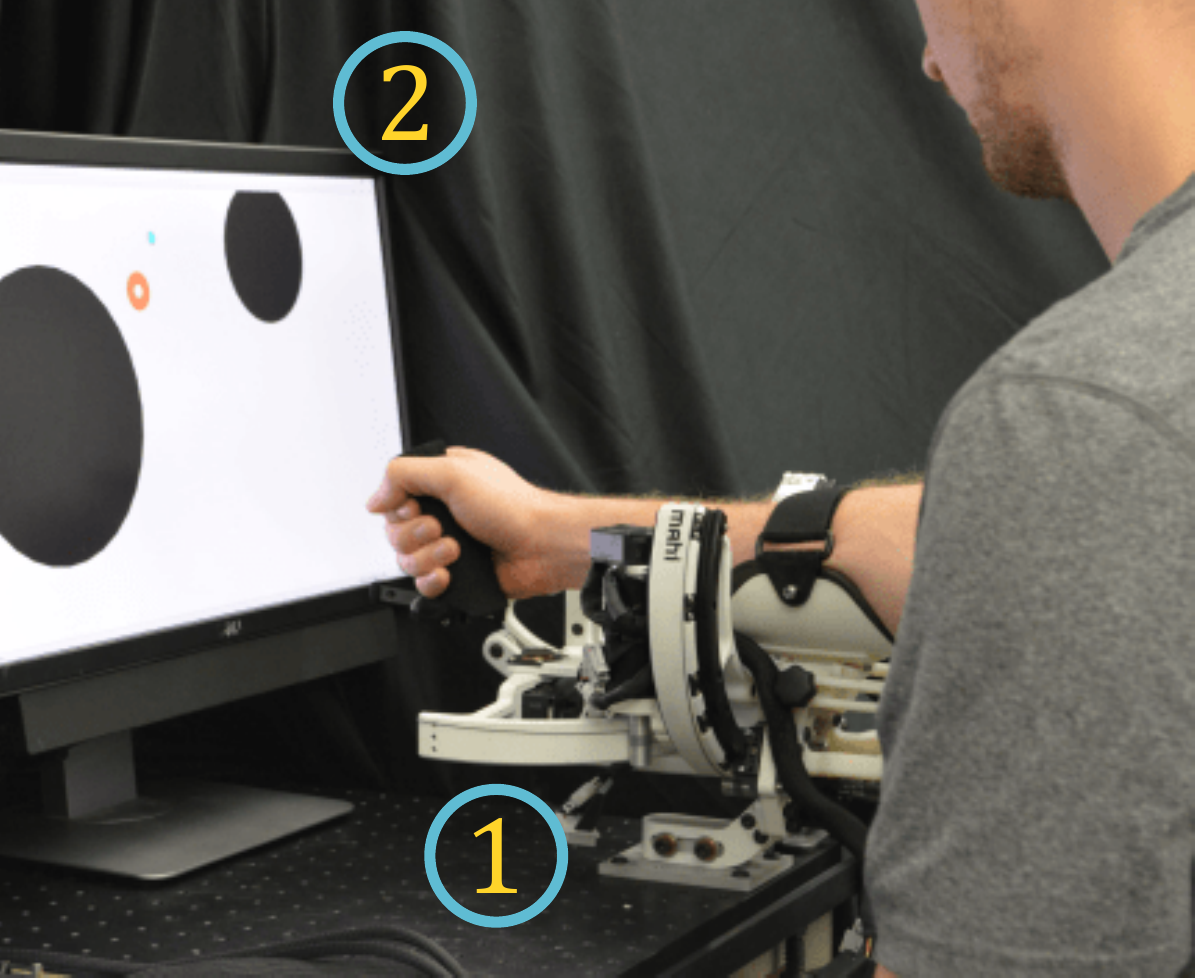}

		\caption{Experimental setup.  (1) participants grasped the handle of the Rice OpenWrist, a multi-DoF haptic device.  (2) by interacting with this robot, participants could move their cursor within the virtual task.}

		\label{fig:setup}
	\end{center}

	\vspace{-0.75em}

\end{figure}

	Ten subjects (three females) participated in this experiment after signing a consent form approved by the Rice University Institutional Review Board.  Subjects physically interacted with the flexion/extension (FE) and radial/ulnar (RU) joints of the OpenWrist, a robotic exoskeleton recently developed for wrist rehabilitation \cite{pezent2017}.  Thus, the human and robot shared control over 2-DoF.  In order to measure the human's input torque about both axes of rotation---FE and RU---we employed the nonlinear disturbance observer originally proposed in \cite{chen2000}, and specifically applied to haptic devices by \cite{gupta2011}.  This disturbance observer, along with the IC and IC-PITD algorithms, were implemented upon a desktop PC through Matlab/Simulink (MathWorks) together with the QUARC blockset (Quansar).  The computer communicated with the haptic device using a Q8-USB data acquisition device (Quansar), and the sample period $T$ was set as $T = 10^{-3}$~s. \par

\begin{figure}[t]

	\begin{center}
		\includegraphics[width=0.65\columnwidth]{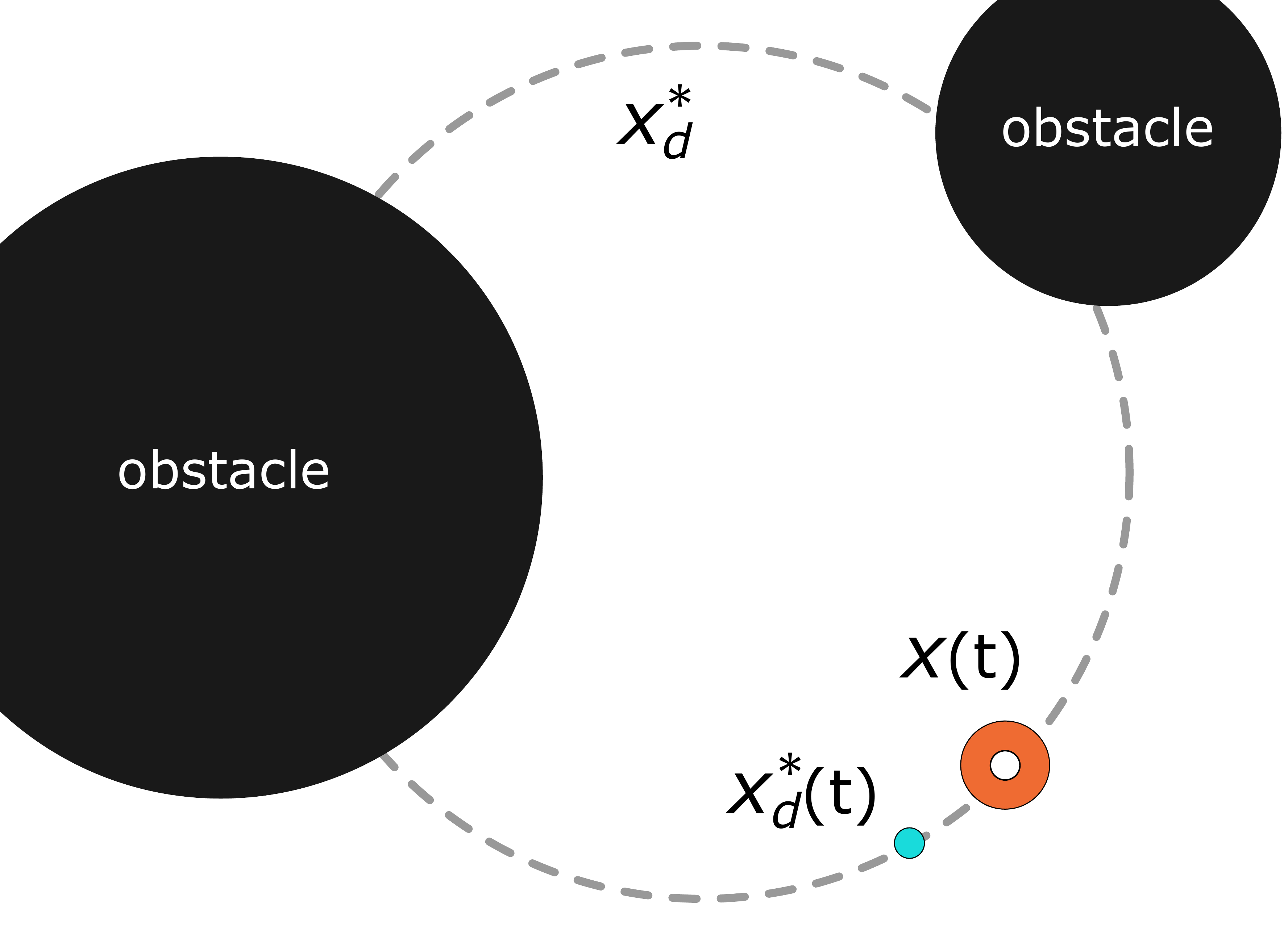}

		\caption{Virtual task.  By rotating the flexion/extension and radial/ulnar joints of the haptic device, subjects changed $\boldsymbol{x}(t)$, their current position (orange torus).  The original desired position at the current time, $\boldsymbol{x_d^*}(t)$, was denoted by the blue sphere.  Without any human intervention, $\boldsymbol{x}(t)$ tracks $\boldsymbol{x_d^*}(t)$ around a circular trajectory (dashed line), and intersects both obstacles.  Participants were asked to interact with the robot such that the torus avoids each obstacle.}

		\label{fig:environment}
	\end{center}

	\vspace{-2em}

\end{figure}

	Subjects observed a computer monitor (see Fig.~\ref{fig:setup}), where the virtual task was displayed.  An image and explanation of this virtual task is provided by Fig.~\ref{fig:environment}.  The robot's original desired trajectory, $\boldsymbol{x_d^*}$, was defined as a circle in joint space
\begin{equation} \label{eq:S7E1}
	\boldsymbol{x_d^*}(t) = \pi/9 \cdot [cos(t), sin(t)]^T
\end{equation}
The robot's position at time $t$, i.e., $\boldsymbol{x}(t)$, was denoted by an orange torus.  Subjects were instructed to move this torus to follow (\ref{eq:S7E1}), while also interacting with the robot to avoid two obstacles.  A single trial consisted of a complete revolution around $\boldsymbol{x_d^*}$.  It should be noted that---because of the IC and IC-PITD algorithms---this was a shared control task, where the robot could autonomously track  (\ref{eq:S7E1}) if no human inputs were present.  The virtual stiffness, $K_d \in \mathbb{R}^2$, and virtual damping, $B_d \in \mathbb{R}^2$, were the same for both IC and IC-PITD.  Based on our prior experience and preliminary experiments, we heuristically selected $K_d = \mbox{diag}(k_d, k_d)$, with $k_d = 35$~Nm/rad, and $B_d = \mbox{diag}(b_d,b_d)$, with $b_d = 0.5$~Nm$\cdot$s/rad.  Recalling our findings from Section~\ref{sec:simulations} {concerning computational efficiency}, we additionally chose $\delta = 10^{-3}$~s, noting that, as a result of $T$ and $\delta$, $r=1$ for this experiment.  {The IC-PITD parameters $\tau$ and $\mu$ were identified offline during preliminary experiments, where we iteratively increased $\tau$ and $\mu$ until we obtained interaction behavior that was noticeably different from impedance control. Using this heuristic process, we selected $\tau = 1.25$~s and $\mu = 0.35$~rad/Nm$\cdot$s.} \par

	We utilized a within-subjects design, where all ten participants completed the same experimental protocol twice: once with IC, and once using IC-PITD.  To better eliminate ordering effects, we counterbalanced which controller was tested first.  Given a controller---either IC or IC-PITD---the experimenter initially demonstrated the virtual task for the subject.  Once that subject understood the task, he or she then practiced for approximately five minutes by performing $50$ unrecorded trials (revolutions).  This training period was meant to familiarize the participant with the current controller, and also to help alleviate any learning effects or differences within initial skill \cite{dragan2013}.  After the training phase, subjects executed $15$ more trials, the last $10$ of which were recorded for our data analysis.  Next, the same protocol---demonstrating, training, and recording trials---was repeated for the second controller.  Accordingly, every participant performed a total of $20$ recorded trials, i.e., $10$ with both controller types (IC and IC-PITD). \par

\subsection{Data Analysis}

	For each recorded trial we computed the (a) applied torque, (b) interaction percent, (c) collision percent, (d) tracking error, and (e) movement smoothness.  \textit{Applied torque} was defined as $\int_0^{2\pi} \|\boldsymbol{f_h}(t)\|dt$, where $\boldsymbol{f_h}(t) \in \mathbb{R}^2$ is the human's torque about the FE and RU axes at time $t$, as estimated by the disturbance observer.  \textit{Interaction percent} indicates to the fraction of the trial during which the magnitude of the human's applied torque exceeded $0.5$~Nm; in other words, $\frac{1}{2\pi} \int_0^{2\pi} \big(\|\boldsymbol{f_h}(t)\| > 0.5\big) dt$.  We use applied torque and interaction percent to assess not only the subject's total effort, but also how persistently the human interacts with the robot.  \textit{Collision percent} refers to the fraction of the trial during which the human's torus is in contact with an obstacle.  \textit{Tracking error} considers the difference between $\boldsymbol{x}(t)$ and $\boldsymbol{x_d^*}(t)$, but only when $\boldsymbol{x_d^*}(t)$ is more than $0.2$~radians away from both of the obstacles.  Let us refer to the segments of time where $\boldsymbol{x_d^*}(t)$ is in this free space as $\tau_{free} \subseteq 2\pi$; then, the tracking error is calculated as $\int_{\tau_{free}} \|x(t) - x_d^*(t)\| dt$.  We use both collision percent and tracking error to determine how accurately the participant completed our virtual task.  Finally, \textit{movement smoothness} was measured by leveraging the spectral arc-length metric \cite{balasubramanian2012}.  Within this spectral-arc length algorithm, an amplitude threshold of $0.05$ and a maximum cut-off frequency of $10$ Hz were selected.  We use movement smoothness, as measured by spectral-arc length, to investigate the human-like quality of the robot's movements. \par

	We separately computed each subject's mean trial metrics across the $10$ recorded trials with IC, as well as their mean trial metrics across the $10$ recorded trials using IC-PITD.  As such, for an individual subject, we obtained their average torque, interaction, collision, error, and smoothness metrics per trial when IC was present, along with the same averaged behavior per trial when IC-PITD was employed.  So as to statistically test the effects of controller type, we then used paired t-tests with a significance level of $\alpha = 0.05$.  Analysis concerning statistical significance was conducted using SPSS (IBM). \par

\subsection{Results and Discussion}

\begin{figure*}[t!]

	\begin{center}
		\includegraphics[width=2\columnwidth]{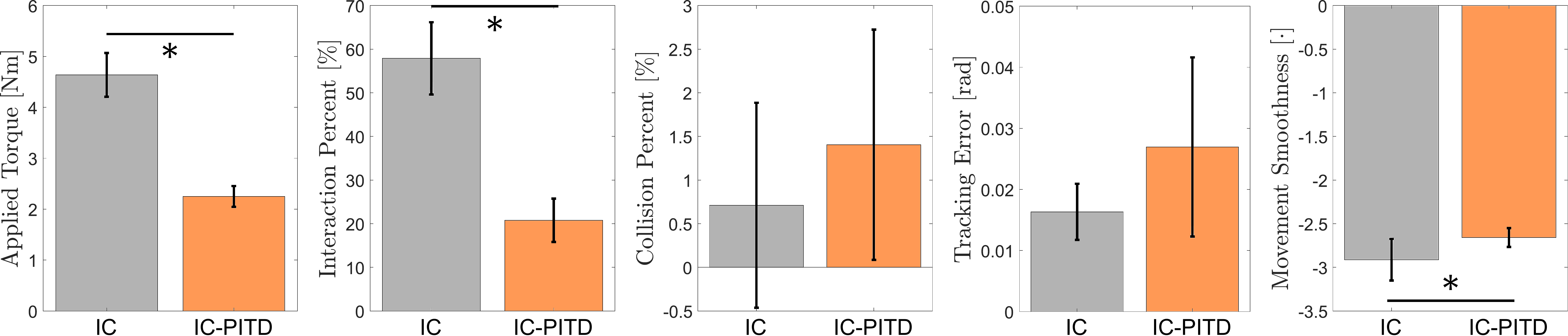}

		\caption{Experimental results.  Each plot shows the mean across subjects when impedance control was used (IC, gray), and when impedance control with physically interactive trajectory deformations was used (IC-PITD, orange).  Error bars indicate the standard deviation, and an asterisk ($*$) denotes $p < 0.05$.}

		\label{fig:stats}
	\end{center}

	\vspace{-2em}

\end{figure*}

	Our findings are summarized by Fig.~\ref{fig:stats}, and reported below in the form $mean \pm std$.  For trials with IC, subjects applied an average torque of $4.64 \pm 0.43$ Nm, while the same subjects applied an average torque of $2.25 \pm 0.20$~Nm during IC-PITD trials.  Similarly, the interaction percent for the IC case, $57.90 \pm 8.29\%$, exceeded the interaction percent with IC-PITD, $20.77 \pm 4.97\%$.  We concluded that there was a statistically significant decrease in both applied torque ($t(9) = 15.907$, $p < 0.001$) and interaction percent ($t(9) = 11.530$, $p < 0.001$) when subjects performed our pHRI task using IC-PITD.  In terms of collision percent, we observed a slight increase between the IC case, $0.71 \pm 1.17\%$, and the IC-PITD trials, $1.41 \pm 1.32\%$, but this difference was not statistically significant ($t(9) = -1.203$, $p = 0.260$).  Interestingly, a similar pattern was discovered for tracking error; in the IC trials, subjects had an average tracking error of $0.016 \pm 0.005$~rad, which was actually lower than the average tracking error of $0.027 \pm 0.015$~rad within the IC-PITD trials.  This increase in error, however, was not determined to be statistically significant ($t(9) = -2.025$, $p = 0.073$).  Finally, subjects had an average movement smoothness of $-2.91 \pm 0.24$ for the IC case and $-2.66 \pm 0.11$ during the IC-PITD trials.  Recalling that---within the spectral arc-length metric---more negative values indicate less smooth movements \cite{balasubramanian2012}, we found that IC-PITD lead to a statistically significant improvement in movement smoothness ($t(9) = -2.640$, $p = 0.027$). \par

	Observing the low collision percent and tracking error, it is clear that subjects were able to successfully avoid the virtual obstacles and then return to following $\boldsymbol{x_d^*}$ while using either IC or IC-PITD.  On the other hand, controller type did have a statistically significant effect on the subjects' efficiency; when physically interactive trajectory deformations was combined with the impedance controller, participants could complete the task while applying less total torque, and did not need to physically interact with the robot as frequently.  Indeed, the robot's resultant movements with IC-PITD were found to be smoother, and thus natural and human-like.  The results from the described experiment therefore support our hypotheses: pHRI with IC-PITD was more efficient and smooth, without noticeably reducing accuracy as compared to IC.  It must be recognized, however, that the same behaviors may not necessarily occur during other tasks, or for individual users with different levels of proficiency.  As such, the results from our comparison of IC and IC-PITD should be understood as practical trends, and not absolute truths.  {Interestingly, based on the informal feedback we received from participants after the experiment, we recommend tuning the impedance controller to be more stiff when implementing IC-PITD.  Subjects might have informally preferred higher $K_d$ gains because IC-PITD provides two sources of compliance---both the impedance controller and the trajectory deformations---and so the stiffness of the impedance controller should be increased in compensation.} \par

\section{Conclusion} \label{sec:conclusion}

	We proposed an impedance control algorithm which allows humans to simultaneously deform the robot's actual and desired trajectories during pHRI.  Under our approach, the human utilizes physical interactions to alter \textit{how} the robot completes some collaborative task.  Since the deformed desired trajectory smoothly returns to the original desired trajectory in the absence of human interactions, the robot maintains a level of autonomy, and both human and robot can meaningfully contribute towards collaborative movements.  We first considered a single trajectory deformation, which was defined by selecting the variation; in particular, endpoint constraints on this variation were identified to prevent interference with impedance control.  Next, the trajectory deformation's energy was written as a balance of the {work done to the human} and the variation's total jerk.  After applying the method of Lagrange multipliers, we found the optimal variation shape which, in practice, could additionally be made invariant to the sampling period.  We then introduced a real time algorithm that combined impedance control with physically interactive trajectory deformations, where the human now continuously deforms the robot's desired trajectory by applying wrenches at the robot's end-effector.  This algorithm is intended for multi-DoF {robots}, and scales linearly with the dimensionality of the task space.  To verify our algorithm, and intuitively understand the effects of its user-specified arbitration parameters, we performed 1-DoF simulations.  Finally, we experimentally demonstrated on human subjects and a 2-DoF haptic device that impedance control with physically interactive trajectory deformations yields efficient and smooth pHRI. \par

	{Future work includes comparing the deformation method presented in this paper to learning approaches that re-plan the robot's desired trajectory.  For instance, we are interested in explicitly learning the human's energy function from pHRI---at each time step where our understanding of the human's energy function is updated, we would then re-plan the robot's desired trajectory to minimize that energy function.  On the one hand, this learning and re-planning approach could lead to less pHRI and a desired trajectory that better matches the human's preferences.  On the other hand, our deformation approach may be more intuitive, and can be performed faster than re-planning the entire trajectory.}  {Another topic for future work concerns the speed at which the task is completed; we recognize that a limitation of our current work is that the human cannot use pHRI to change the robot's overall task timing.  In future work, we would like to extend learning methods in order to infer the human's desired timing through pHRI.} \par

\bibliographystyle{IEEEtran}
\bibliography{IEEEabrv,BibFile}

\begin{thebibliography}{10}
\providecommand{\url}[1]{#1}
\csname url@samestyle\endcsname
\providecommand{\newblock}{\relax}
\providecommand{\bibinfo}[2]{#2}
\providecommand{\BIBentrySTDinterwordspacing}{\spaceskip=0pt\relax}
\providecommand{\BIBentryALTinterwordstretchfactor}{4}
\providecommand{\BIBentryALTinterwordspacing}{\spaceskip=\fontdimen2\font plus
\BIBentryALTinterwordstretchfactor\fontdimen3\font minus
  \fontdimen4\font\relax}
\providecommand{\BIBforeignlanguage}[2]{{%
\expandafter\ifx\csname l@#1\endcsname\relax
\typeout{** WARNING: IEEEtran.bst: No hyphenation pattern has been}%
\typeout{** loaded for the language `#1'. Using the pattern for}%
\typeout{** the default language instead.}%
\else
\language=\csname l@#1\endcsname
\fi
#2}}
\providecommand{\BIBdecl}{\relax}
\BIBdecl

\bibitem{jarrasse2012}
N.~Jarrass{\'e}, T.~Charalambous, and E.~Burdet, ``A framework to describe,
  analyze and generate interactive motor behaviors,'' \emph{PLOS ONE}, vol.~7,
  no.~11, p. e49945, 2012.

\bibitem{bratman1992}
M.~E. Bratman, ``Shared cooperative activity,'' \emph{Phil. Review}, vol. 101,
  no.~2, pp. 327--341, 1992.

\bibitem{hogan1985}
N.~Hogan, ``Impedance control: An approach to manipulation,'' \emph{J. Dyn.
  Syst. Meas. Control}, vol. 107, no.~1, pp. 1--24, 1985.

\bibitem{santis2008}
A.~De~Santis, B.~Siciliano, A.~De~Luca, and A.~Bicchi, ``An atlas of physical
  human--robot interaction,'' \emph{Mechanism and Machine Theory}, vol.~43,
  no.~3, pp. 253--270, 2008.

\bibitem{haddadin2016}
S.~Haddadin and E.~Croft, ``Physical human-robot interaction,'' in
  \emph{Springer Handbook of Robotics}, 2nd~ed., 2016, pp. 1835--1874.

\bibitem{jarrasse2008}
N.~Jarrass{\'e}, J.~Paik, V.~Pasqui, and G.~Morel, ``How can human motion
  prediction increase transparency?'' in \emph{Proc. IEEE Conf. Robot. Autom.},
  2008, pp. 2134--2139.

\bibitem{corteville2007}
B.~Corteville, E.~Aertbeli{\"e}n, H.~Bruyninckx, J.~De~Schutter, and
  H.~Van~Brussel, ``Human-inspired robot assistant for fast point-to-point
  movements,'' in \emph{Proc. IEEE Conf. Robot. Autom.}, 2007, pp. 3639--3644.

\bibitem{li2014}
Y.~Li and S.~S. Ge, ``Human--robot collaboration based on motion intention
  estimation,'' \emph{IEEE/ASME Trans. Mechatronics}, vol.~19, no.~3, pp.
  1007--1014, 2014.

\bibitem{erden2010}
M.~S. Erden and T.~Tomiyama, ``Human-intent detection and physically
  interactive control of a robot without force sensors,'' \emph{IEEE Trans.
  Robot.}, vol.~26, no.~2, pp. 370--382, 2010.

\bibitem{mortl2012}
A.~M{\"o}rtl, M.~Lawitzky, A.~Kucukyilmaz, M.~Sezgin, C.~Basdogan, and
  S.~Hirche, ``The role of roles: {P}hysical cooperation between humans and
  robots,'' \emph{Int. J. Robot. Res.}, vol.~31, no.~13, pp. 1656--1674, 2012.

\bibitem{li2015}
Y.~Li, K.~P. Tee, W.~L. Chan, R.~Yan, Y.~Chua, and D.~K. Limbu, ``Continuous
  role adaptation for human--robot shared control,'' \emph{IEEE Trans. Robot.},
  vol.~31, no.~3, pp. 672--681, 2015.

\bibitem{kucukyilmaz2013}
A.~Kucukyilmaz, T.~M. Sezgin, and C.~Basdogan, ``Intention recognition for
  dynamic role exchange in haptic collaboration,'' \emph{IEEE Trans. Haptics},
  vol.~6, no.~1, pp. 58--68, 2013.

\bibitem{medina2015}
J.~R. Medina, T.~Lorenz, and S.~Hirche, ``Synthesizing anticipatory haptic
  assistance considering human behavior uncertainty,'' \emph{IEEE Trans.
  Robot.}, vol.~31, no.~1, pp. 180--190, 2015.

\bibitem{peternel2016}
L.~Peternel, N.~Tsagarakis, and A.~Ajoudani, ``Towards multi-modal intention
  interfaces for human-robot co-manipulation,'' in \emph{Proc. IEEE/RSJ Conf.
  Intelligent Robots Syst.}, 2016, pp. 2663--2669.

\bibitem{masone2012}
C.~Masone, A.~Franchi, H.~H. B{\"u}lthoff, and P.~R. Giordano, ``Interactive
  planning of persistent trajectories for human-assisted navigation of mobile
  robots,'' in \emph{Proc. IEEE/RSJ Conf. Intelligent Robots Syst.}, 2012, pp.
  2641--2648.

\bibitem{masone2014}
C.~Masone, P.~R. Giordano, H.~H. B{\"u}lthoff, and A.~Franchi,
  ``Semi-autonomous trajectory generation for mobile robots with integral
  haptic shared control,'' in \emph{Proc. IEEE Conf. Robot. Autom.}, 2014, pp.
  6468--6475.

\bibitem{mainprice2013}
J.~Mainprice and D.~Berenson, ``Human-robot collaborative manipulation planning
  using early prediction of human motion,'' in \emph{Proc. IEEE/RSJ Conf.
  Intelligent Robots Syst.}, 2013, pp. 299--306.

\bibitem{chao2016}
C.~Chao and A.~Thomaz, ``Timed petri nets for fluent turn-taking over
  multimodal interaction resources in human-robot collaboration,''
  \emph{‎Int. J. Robot. Res}, vol.~35, no.~11, pp. 1330 -- 1353, 2016.

\bibitem{lasota2015}
P.~A. Lasota and J.~A. Shah, ``Analyzing the effects of human-aware motion
  planning on close-proximity human-robot collaboration,'' \emph{Hum. Factors},
  vol.~57, no.~1, pp. 21--33, 2015.

\bibitem{pham2015}
Q.-C. Pham and Y.~Nakamura, ``A new trajectory deformation algorithm based on
  affine transformations,'' \emph{IEEE Trans. Robot.}, vol.~31, no.~4, pp.
  1054--1063, 2015.

\bibitem{brock2002}
O.~Brock and O.~Khatib, ``Elastic strips: A framework for motion generation in
  human environments,'' \emph{‎Int. J. Robot. Res}, vol.~21, no.~12, pp.
  1031--1052, 2002.

\bibitem{zucker2013}
M.~Zucker, N.~Ratliff, A.~D. Dragan, M.~Pivtoraiko, M.~Klingensmith, C.~M.
  Dellin, J.~A. Bagnell, and S.~S. Srinivasa, ``{CHOMP}: {C}ovariant
  {H}amiltonian optimization for motion planning,'' \emph{‎Int. J. Robot.
  Res}, vol.~32, no. 9-10, pp. 1164--1193, 2013.

\bibitem{kalakrishnan2011}
M.~Kalakrishnan, S.~Chitta, E.~Theodorou, P.~Pastor, and S.~Schaal, ``{STOMP}:
  {S}tochastic trajectory optimization for motion planning,'' in \emph{Proc.
  IEEE Conf. Robot. Autom.}, 2011, pp. 4569--4574.

\bibitem{schulman2014}
J.~Schulman, Y.~Duan, J.~Ho, A.~Lee, I.~Awwal, H.~Bradlow, J.~Pan, S.~Patil,
  K.~Goldberg, and P.~Abbeel, ``Motion planning with sequential convex
  optimization and convex collision checking,'' \emph{‎Int. J. Robot. Res},
  vol.~33, no.~9, pp. 1251--1270, 2014.

\bibitem{park2012}
C.~Park, J.~Pan, and D.~Manocha, ``{ITOMP}: {I}ncremental trajectory
  optimization for real-time replanning in dynamic environments.'' in
  \emph{Proc. Conf. Autom. Planning Scheduling}, 2012, pp. 207--215.

\bibitem{dragan2013}
A.~D. Dragan and S.~S. Srinivasa, ``A policy-blending formalism for shared
  control,'' \emph{Int. J. Robot. Res.}, vol.~32, no.~7, pp. 790--805, 2013.

\bibitem{diolaiti2006}
N.~Diolaiti, G.~Niemeyer, F.~Barbagli, and J.~K. Salisbury, ``Stability of
  haptic rendering: {D}iscretization, quantization, time delay, and coulomb
  effects,'' \emph{IEEE Trans. Robot.}, vol.~22, no.~2, pp. 256--268, 2006.

\bibitem{abbott2005}
J.~J. Abbott and A.~M. Okamura, ``Effects of position quantization and sampling
  rate on virtual-wall passivity,'' \emph{IEEE Trans. Robot.}, vol.~21, no.~5,
  pp. 952--964, 2005.

\bibitem{colgate1994}
J.~E. Colgate and J.~M. Brown, ``Factors affecting the {Z}-width of a haptic
  display,'' in \emph{Proc. IEEE Conf. Robot. Autom.}, 1994, pp. 3205--3210.

\bibitem{lee2006}
J.~M. Lee, \emph{Riemannian manifolds: {A}n introduction to curvature}.\hskip
  1em plus 0.5em minus 0.4em\relax New York, NY, USA: Springer Science \&
  Business Media, 2006.

\bibitem{reed2008}
K.~B. Reed and M.~A. Peshkin, ``Physical collaboration of human-human and
  human-robot teams,'' \emph{IEEE Trans. Haptics}, vol.~1, no.~2, pp. 108--120,
  2008.

\bibitem{noohi2016}
E.~Noohi, M.~{\v{Z}}efran, and J.~L. Patton, ``A model for human--human
  collaborative object manipulation and its application to human--robot
  interaction,'' \emph{IEEE Trans. Robot.}, vol.~32, no.~4, pp. 880--896, 2016.

\bibitem{rao2009}
S.~S. Rao, \emph{Engineering optimization: {T}heory and practice},
  4th~ed.\hskip 1em plus 0.5em minus 0.4em\relax New York, NY, USA: John Wiley
  \& Sons, 2009.

\bibitem{chipalkatty2013}
R.~Chipalkatty, G.~Droge, and M.~B. Egerstedt, ``Less is more:
  {M}ixed-initiative model-predictive control with human inputs,'' \emph{IEEE
  Trans. Robot.}, vol.~29, no.~3, pp. 695--703, 2013.

\bibitem{flash1985}
T.~Flash and N.~Hogan, ``The coordination of arm movements: {A}n experimentally
  confirmed mathematical model,'' \emph{J. Neurosci.}, vol.~5, no.~7, pp.
  1688--1703, 1985.

\bibitem{meirovitch2016}
Y.~Meirovitch, D.~Bennequin, and T.~Flash, ``Geometrical invariance and
  smoothness maximization for task-space movement generation,'' \emph{IEEE
  Trans. Robot.}, vol.~32, no.~4, pp. 837--853, 2016.

\bibitem{dragan2015}
A.~D. Dragan, ``Legible robot motion planning,'' Ph.D. dissertation, Robotics
  Institute, CMU, Pittsburgh, PA, 2015.

\bibitem{spong2006}
M.~W. Spong, S.~Hutchinson, and M.~Vidyasagar, \emph{Robot modeling and
  control}.\hskip 1em plus 0.5em minus 0.4em\relax New York, NY, USA: John
  Wiley \& Sons, 2006, vol.~3.

\bibitem{pezent2017}
E.~Pezent, C.~G. Rose, A.~D. Deshpande, and M.~K. O'Malley, ``Design and
  characterization of the openwrist: A robotic wrist exoskeleton for
  coordinated hand-wrist rehabilitation,'' in \emph{Proc. IEEE Conf. Rehab.
  Robot.}, 2017, pp. 720--725.

\bibitem{chen2000}
W.-H. Chen, D.~J. Ballance, P.~J. Gawthrop, and J.~O'Reilly, ``A nonlinear
  disturbance observer for robotic manipulators,'' \emph{IEEE Trans. Ind.
  Electron.}, vol.~47, no.~4, pp. 932--938, 2000.

\bibitem{gupta2011}
A.~Gupta and M.~K. O'Malley, ``Disturbance-observer-based force estimation for
  haptic feedback,'' \emph{J. Dyn. Syst. Meas. Control}, vol. 133, no.~1, p.
  014505, 2011.

\bibitem{balasubramanian2012}
S.~Balasubramanian, A.~Melendez-Calderon, and E.~Burdet, ``A robust and
  sensitive metric for quantifying movement smoothness,'' \emph{IEEE Trans.
  Biomed. Eng.}, vol.~59, no.~8, pp. 2126--2136, 2012.

\end{thebibliography}

\begin{biography}
[{\includegraphics[width=1in,height=1.25in,clip,keepaspectratio]{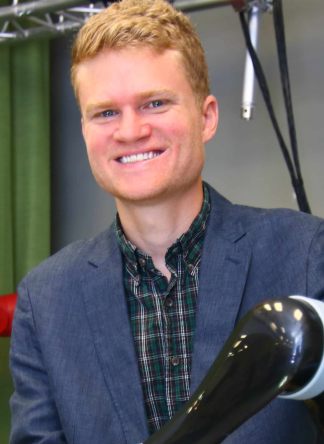}}]
{Dylan P. Losey}
(S'14) received the B.S. degree in mechanical engineering from Vanderbilt University, Nashville, TN, USA, in 2014, and the M.S. degree in mechanical engineering from Rice University, Houston, TX, USA, in 2016. \par
He is currently working towards the Ph.D. degree in mechanical engineering at Rice Univserity, where he has been a member of the Mechatronics and Haptic Interfaces Laboratory since 2014.  In addition, between May and August 2017, he was a visiting scholar in the Interactive Autonomy and Collaborative Technologies Laboratory at the University of California, Berkeley.  He researches physical human-robot interaction; in particular, how robots can learn from and adapt to human corrections. \par
Mr. Losey received an NSF Graduate Research Fellowship in 2014, and the 2016 IEEE/ASME Transactions on Mechatronics Best Paper Award.
\end{biography}

\begin{biography}[{\includegraphics[width=1in,height=1.25in,clip,keepaspectratio]{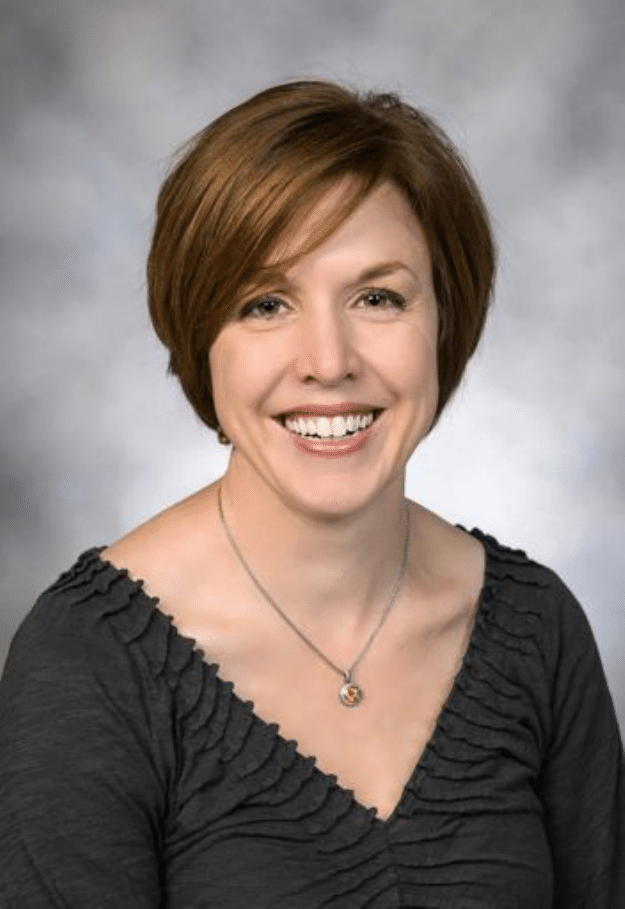}}]
{Marcia K. O'Malley}
(SM'13) received the B.S. degree in mechanical engineering from Purdue University, West Lafayette, IN, USA, in 1996, and the M.S. and Ph.D. degrees in mechanical engineering from Vanderbilt University, Nashville, TN, USA, in 1999 and 2001, respectively. \par
She is currently a Professor of mechanical engineering and of computer science and of electrical and computer engineering at Rice University, Houston, TX, USA, and directs the Mechatronics and Haptic Interfaces Laboratory. She is adjunct faculty in the Departments of Physical Medicine and Rehabilitation at the Baylor College of Medicine and the University of Texas Medical School at Houston, and is the Director of Rehabilitation Engineering at TIRR-Memorial Hermann Hospital. Her research addresses issues that arise when humans physically interact with robotic systems, with a focus on training and rehabilitation in virtual environments. \par
Prof. O'Malley is a Fellow of the American Society of Mechanical Engineers and serves as an Associate Editor for the ASME Journal of Mechanisms and Robotics, the ACM Transactions on Human Computer Interaction, and the IEEE Transactions on Robotics.
\end{biography}

\end{document}